\author{Paul Kishan Rubenstein}
\date{\today}
\documentclass{article}
\usepackage{microtype}
\usepackage{graphicx}
\usepackage{graphics}
\usepackage{subcaption}
\usepackage{booktabs} % for professional tables
\usepackage[utf8]{inputenc}
\usepackage{tikz}
\usepackage{rotating}

\def\X{\mathcal{X}}

\def\E{\mathbb{E}}
\def\dZ{d_{\mathcal{Z}}}
\def\dX{d_{\mathcal{X}}}
\def\dP{d_{\mathcal{I}}}
\def\Z{\mathcal{Z}}
\def\R{\mathbb{R}}

\def\KL{\mathrm{KL}}

\def\dI{d_\mathcal{I}}

\usepackage{hyperref}
\usepackage{enumitem}

\usepackage{amsmath,amsfonts, amssymb, amsthm}
\usepackage[accepted]{icml2018}

\newlength\figureheight
\newlength\figurewidth

\icmltitlerunning{On the Latent Space of Wasserstein Auto-Encoders}

\begin{document}
\twocolumn[
\icmltitle{On the Latent Space of Wasserstein Auto-Encoders}




\begin{icmlauthorlist}
	\icmlauthor{Paul K. Rubenstein}{mpi,cam}
	\icmlauthor{Bernhard Sch\"olkopf}{mpi}
	\icmlauthor{Ilya Tolstikhin}{mpi}
\end{icmlauthorlist}

\icmlaffiliation{mpi}{Empirical Inference, Max Planck Institute for Intelligent Systems, Tuebingen, Germany}
\icmlaffiliation{cam}{Machine Learning Group, University of Cambridge, United Kingdom}

\icmlcorrespondingauthor{Paul K. Rubenstein}{paul.rubenstein@tuebingen.mpg.de}

\icmlkeywords{Machine Learning, ICML}

\vskip 0.3in
]



\printAffiliationsAndNotice{}  % leave blank if no need to mention equal contribution
\begin{abstract}
We study the role of latent space dimensionality in Wasserstein auto-encoders (WAEs). 
Through experimentation on synthetic and real datasets, we argue that random encoders should be preferred over deterministic encoders.
We highlight the potential of WAEs for representation learning with promising results on a benchmark disentanglement task.
\end{abstract}
\vspace{-0.4cm}
\section{Introduction}
Unsupervised generative modeling is increasingly attracting the attention of the machine learning community.
Given a collection of unlabelled data points $S_X$, the ultimate goal of the task is to tune a model capable of generating sets of synthetic points $S_G$ which \emph{look similar to} $S_X$.
The closely related field of unsupervised representation learning in addition aims to produce semantically meaningful representations (or features) of the data points~$S_X$.

There are various ways of defining the notion of \emph{similarity} between two sets of data points.
The most common approach assumes that both $S_X$ and $S_G$ are sampled independently from two probability distributions $P_X$ and $P_G$ respectively, and employ some of the known divergence measures for distributions.

Two major approaches currently dominate this field.
Variational Auto-Encoders (VAEs) \cite{KW14} minimize the Kullback-Leibler (KL) divergence $D_{\KL}(P_X, P_G)$, which is equivalent to maximizing the \emph{marginal log-likelihood} of the model $P_G$.
Generative Adversarial Networks (GANs) \cite{goodfellow2014generative} employ an elegant framework, commonly referred to as \emph{adversarial training}, which is suitable for many different divergence measures, including (but not limited to) $f$-divergences \cite{nowozin2016f}, 1-Wasserstein distance \cite{AB17}, and Maximum Mean Discrepancy (MMD) \cite{BS+18}.

Both approaches have their pros and cons. 
VAEs can both generate and \emph{encode} (featurize) data points, are stable to train, and typically manage to cover all modes of the data distribution.
Unfortunately, they often produce examples~that are far from the true data manifold.
This is especially true for structured high-dimensional datasets such as natural images, where VAEs generate \emph{blurry}~pictures.

GANs, on the other hand, are good at producing realistic looking examples (landing on or very close to the manifold), however they cannot featurize the points, often cover only few modes of the data distribution, and are highly unstable to train.
A number of recent papers \cite{MSJ+16, MNG17} propose different ways to blend auto-encoding architectures of VAEs with adversarial training in the hope of getting the best of both worlds.

The sample quality of VAEs was recently addressed by Wasserstein Auto-Encoders (WAE) \cite{TBG+17}.
By switching the focus from the KL objective to the optimal transport distance, the authors presented an auto-encoder architecture sharing most of the nice properties of VAEs while providing samples of better quality.
Importantly, WAEs still allow for adversary-free versions, resulting in a min-min formulation leading to stable training.

In this work we aim at further improving the quality of generative modeling and representation learning techniques.
We will focus on adversary-free auto-encoding architectures of WAE, as we find the instability of the adversarial training to be an unfortunate obstacle when it comes to controlled reproducible experiments.

We address some of the important design choices of WAEs related to the properties of the latent space which were not discussed in \cite{TBG+17}.
Based on new theoretical insights, we provide concrete algorithmic suggestions and report empirical results verifying our conclusions.
Our main contributions are:
\vspace{-.35cm}
\begin{enumerate}
\item
We illustrate different ways in which a mismatch between the latent space dimensionality $\dZ$ and the intrinsic data dimensionality $\dP$ may hurt the performance of WAEs (Section \ref{section:fading-squares}).
\item
We argue that WAEs can be made adaptive to the unknown intrinsic data dimensionality $\dP$ by using \emph{probabilistic (random)} encoders rather than the deterministic encoders used in all experiments of \citet{TBG+17}.
The performance of random encoders is on par with the deterministic ones when $d_{\Z} \leq \dP$, and potentially better when $d_{\Z} \gg \dP$ which is typical for real-world applications (Section \ref{subsection:celebA}). 
This suggests that random encoders should generally be preferred when using WAEs.
\item
We verify these conclusions with experiments on synthetic (newly introduced \emph{fading squares}) and real world (CelebA) image datasets.
\item
We evaluate the usefulness of WAEs with random encoders for representation learning by running them on the \emph{dSprites} dataset, a benchmark task in learning ``disentangled representations'' (Section \ref{section:disentanglement}).
WAEs are capable of achieving \emph{simultaneously} better test reconstruction and disentanglement quality as compared to the current state-of-the-art method $\beta$-VAE \cite{HM+17}.
We conclude that WAEs are a promising direction for future research in this field, because compared to VAEs and $\beta$-VAEs they allow more freedom in shaping the learned latent data manifold.
\end{enumerate}

We finish this section with a short description of WAEs, preliminaries, and notations.
\subsection*{Short introduction to Wasserstein auto-encoders}

Similarly to VAEs, WAEs describe a particular way to train probabilistic \emph{latent variable models} (LVMs) $P_G$.
LVMs act by first sampling a code (feature) vector $Z$ from a \emph{prior distribution} $P_Z$ defined over the latent space $\Z$ and then mapping it to a random input point $X\in\X$ using a conditional distribution $P_G(X|Z)$ also known as \emph{the decoder}.
We will be mostly working with image datasets, so for simplicity we set $\X=\R^{\dX}$,\:$\Z=\R^{\dZ}$, and refer to points $x\in\X$ as pictures, images, or inputs interchangeably.

Instead of minimizing the KL divergence between the LVM $P_G$ and the unknown data distribution $P_X$ as done by VAEs, WAEs aim at minimizing any optimal transport distance between them.
Given any non-negative cost function $c(x,x')$ between two images, WAEs minimize the following objective
with respect to parameters of the decoder $P_G(X|Z)$:
\vspace{-.3cm}
\begin{equation}
\label{eq:WAEobj}
\min_{Q(Z|X)}\; \mathop{\E}\limits_{P_X}\; \mathop{\E}\limits_{Q(Z|X)}\bigl[ c\bigl(X, G(Z)\bigr) \bigr] + \lambda \mathcal{D}_Z(Q_Z, P_Z),
\end{equation}
where the conditional distributions $Q(Z|X)$ are commonly known as \emph{encoders}, 
$Q_Z(Z):= \int Q(Z|X) P_X(X) dX$ is \emph{the aggregated posterior} distribution,
$\mathcal{D}_Z$ is any divergence measure between two distributions over $\Z$, and
$\lambda>0$ is a regularization coefficient.
In practice $Q(Z|X=x)$ and $G(z)$ are often parametrized with deep nets, in which case back propagation can be used with stochastic gradient descent techniques to optimize the objective.

We will only consider \emph{deterministic} decoders ${P_G(X|Z=z)}=\delta_{G(z)}$ mapping\footnote{Here $\delta_t$ is a point distribution supported on $t$.} codes $z\in \Z$ to pictures $G(z)\in \X$.
Another design choice that must be made when using a WAE is whether the encoder should map an image $x\in\mathcal{X}$ to a \emph{distribution} $Q(Z|X=x)$ over the latent space or to a single code $z = \varphi(x)\in\mathcal{Z}$,~i.e.\:$Q(Z|X=x) = \delta_{\varphi(x)}$. We refer to the former type as \emph{random} encoders and the latter as \emph{deterministic} encoders.

The objective \eqref{eq:WAEobj} is similar to that of the VAE and has two terms. 
The first \emph{reconstruction term} aligns the encoder-decoder pair so that the encoded images can be accurately reconstructed by the decoder as measured by the cost function $c$ (we will only use the \emph{cross-entropy loss}  throughout).

The second regularization term is different from VAEs: it forces the aggregated posterior $Q_Z$ to match the prior distribution $P_Z$ rather than asking point-wise posteriors $Q(Z|X=x)$ to match $P_Z$ simultaneously for all data points~$x$.
To better understand the difference, note that $Q_Z$ is the distribution obtained by averaging conditional distributions $Q(Z|X=x)$ for all different points $x$ drawn from the data distribution $P_X$.
This means that WAEs explicitly control the shape of the \emph{entire} encoded dataset while VAEs constrain every input point separately.

Both existing versions of the algorithm---WAE-GAN based on adversarial training and the adversary-free WAE-MMD based on the maximum mean discrepancy, only the latter of which we use in this paper---allow for \emph{any} prior distributions $P_Z$ and encoders $Q(Z|X)$ as long as $P_Z$ and $Q_Z$ can be efficiently sampled.
As a result, the WAE model may be easily endowed  with prior knowledge about the possible structure of the dataset through the choice of $P_Z$.

\emph{Notation.} We denote $\varphi(x) = \mathbb{E}\left[Q(Z|X=x)\right]$ to be the mean of the encoding distribution for a given input $x$. By $\varphi_i(x)$ we mean the $i$th coordinate of $\varphi(x)$, $1\leq i \leq \dZ$, and by $P(Z_i)$ and $Q(Z_i)$ the marginal distributions of the prior and aggregated posteriors over the $i$th dimension of $\Z$ respectively.

\section{Dimension mismatch and random encoders}

\begin{figure*}[t!]
	\centering
	\begin{minipage}{.01\textwidth}
		\begin{turn}{90}
			\hspace{1cm} Random encoder \hspace{2.5cm} Deterministic encoder \hspace{1cm}
		\end{turn}       
	\end{minipage}
	\begin{minipage}{.315\textwidth}
		\centering
		(a) What MMD sees
		
		\includegraphics[width=\textwidth]{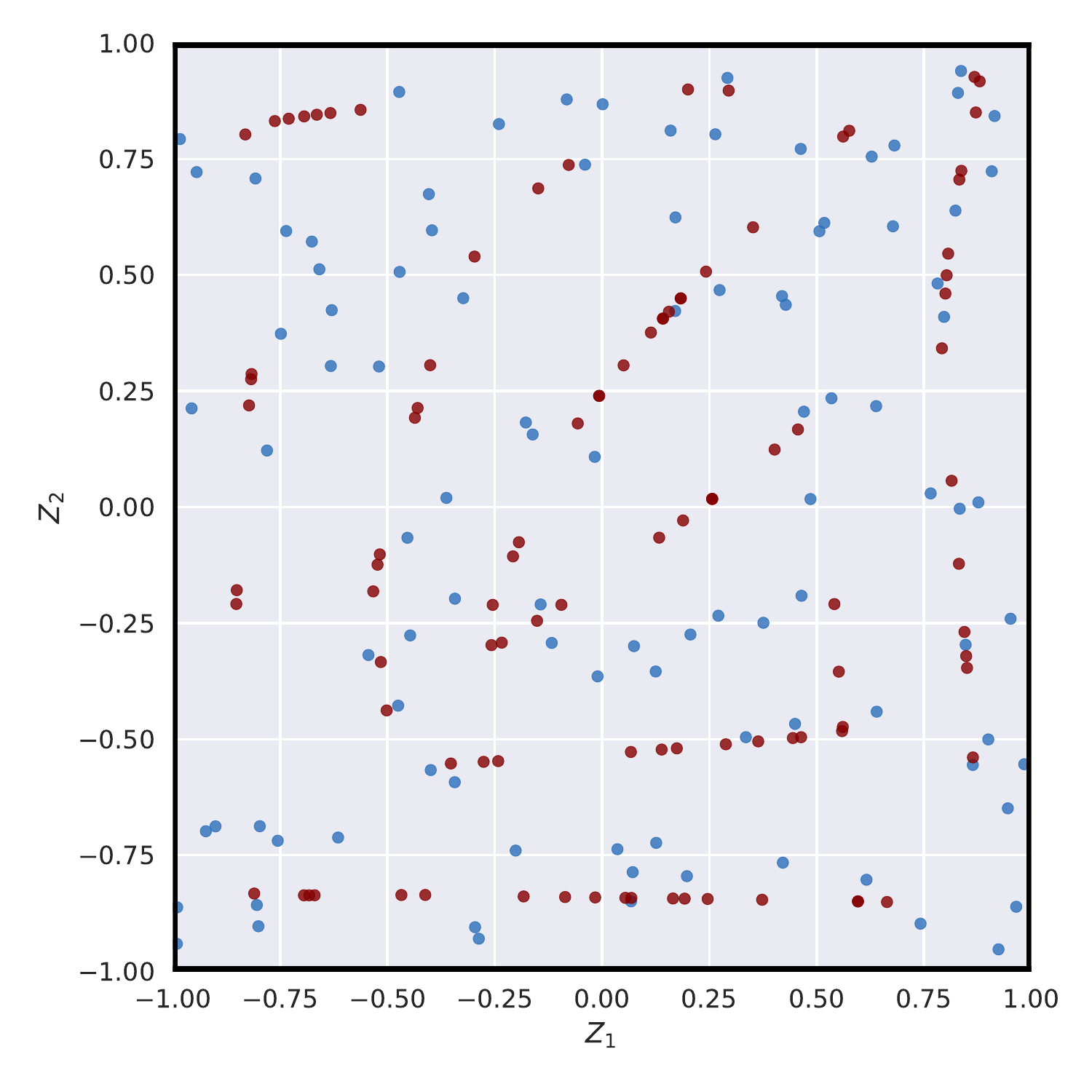}

\includegraphics[width=\textwidth]{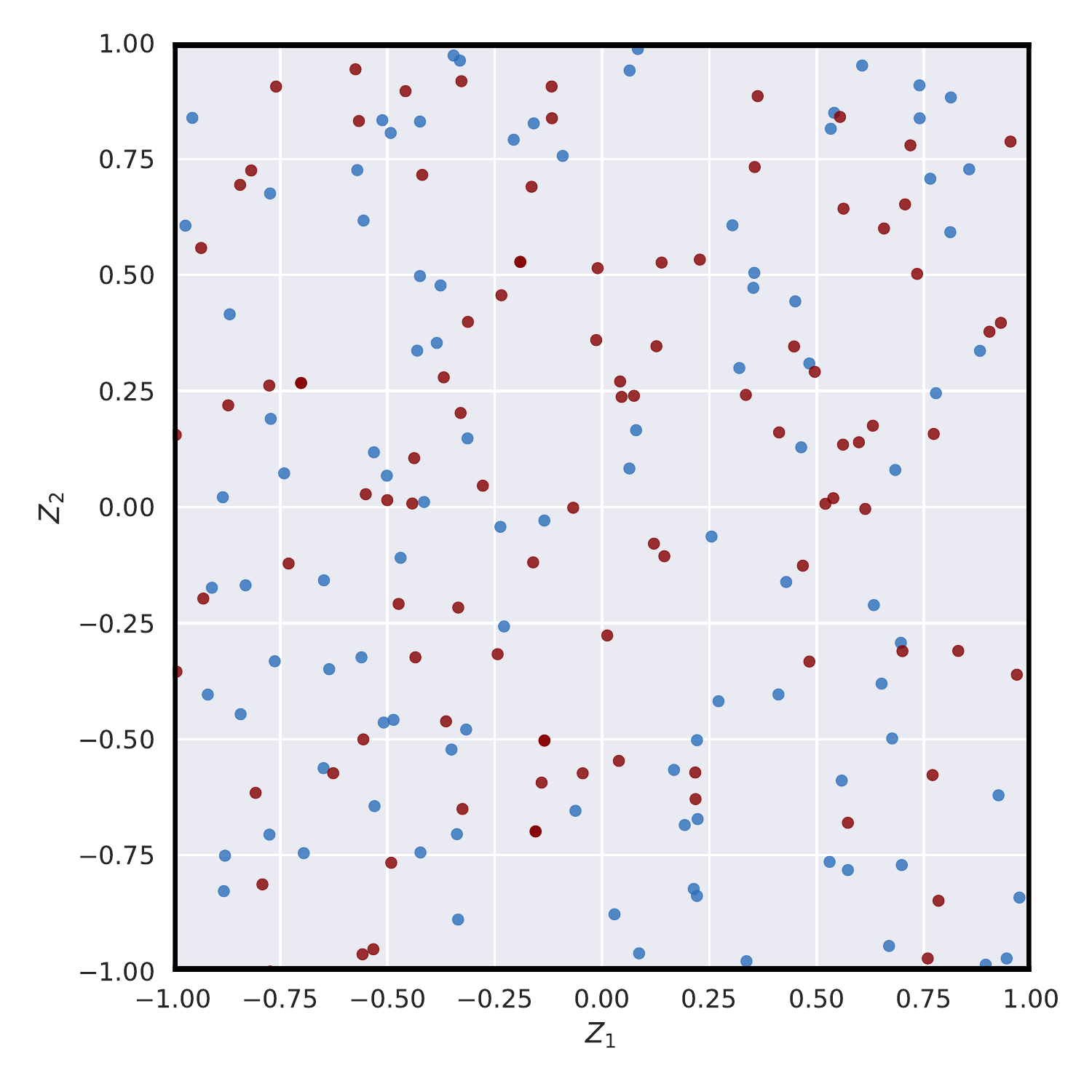}
		
	\end{minipage}%
	\begin{minipage}{.315\textwidth}
		\centering
		(b) What the encoder does
		
		\includegraphics[width=\textwidth]{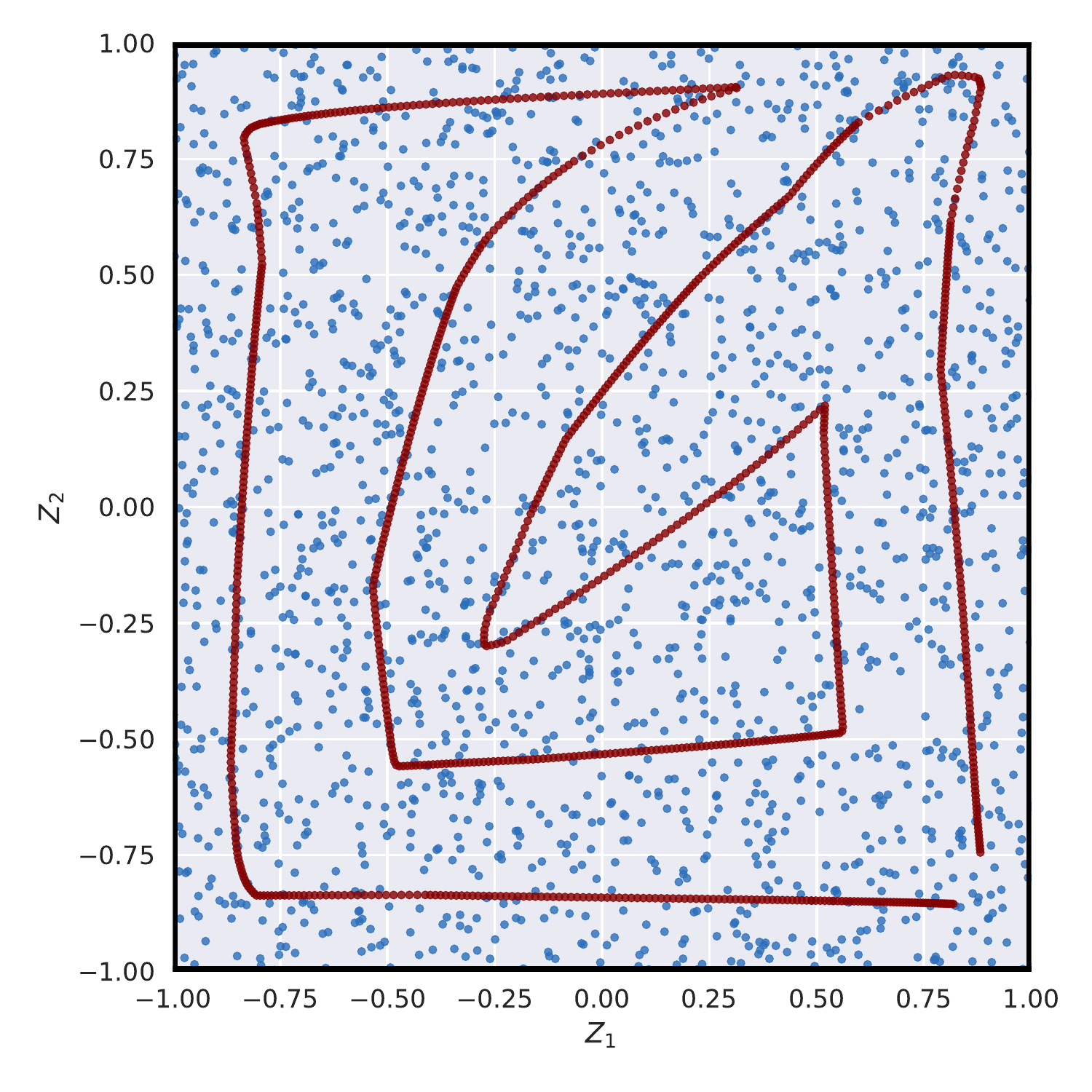}

		\includegraphics[width=\textwidth]{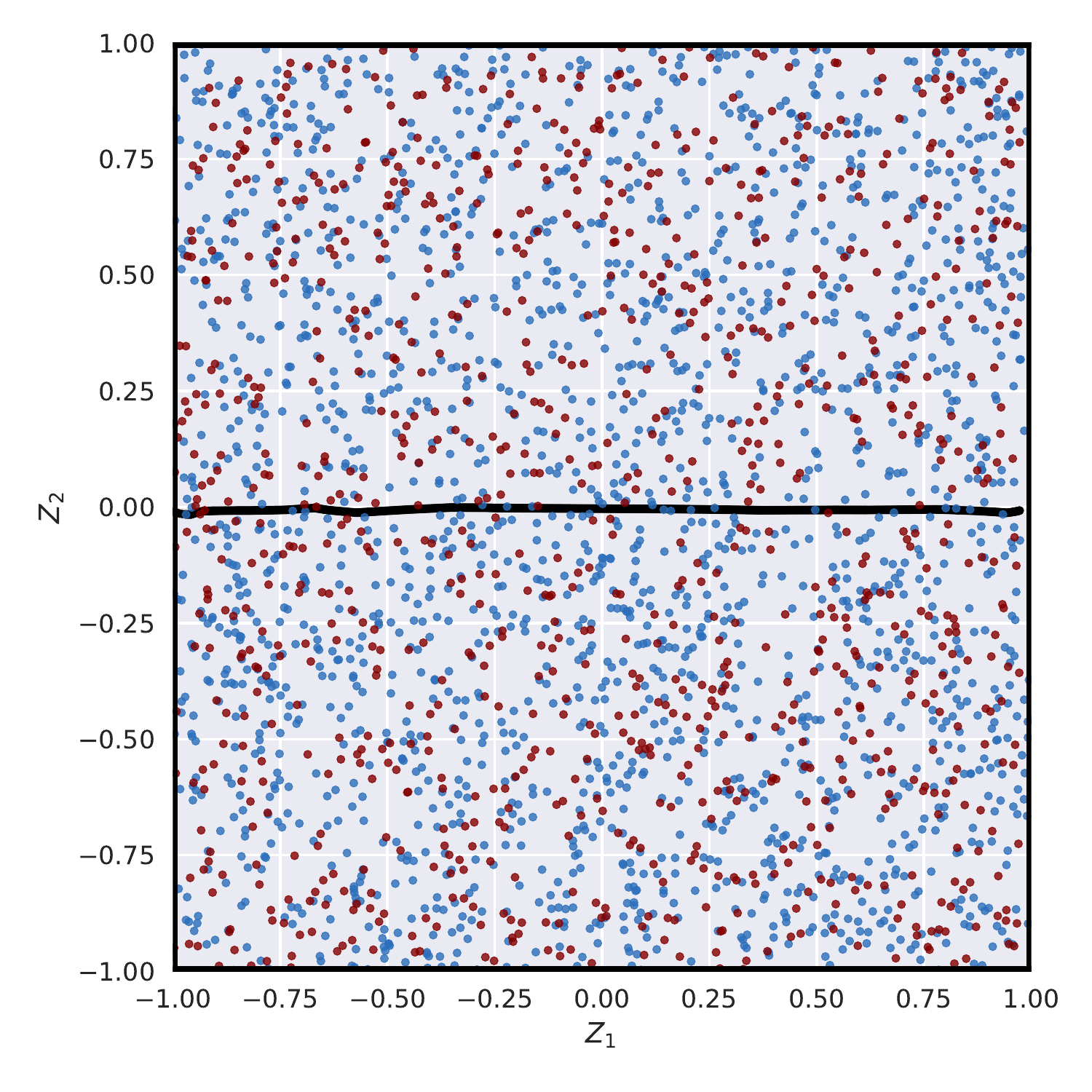}
		
	\end{minipage}%
	\begin{minipage}{.315\textwidth}
		\centering
		(c) What the decoder does
		
		\includegraphics[width=\textwidth]{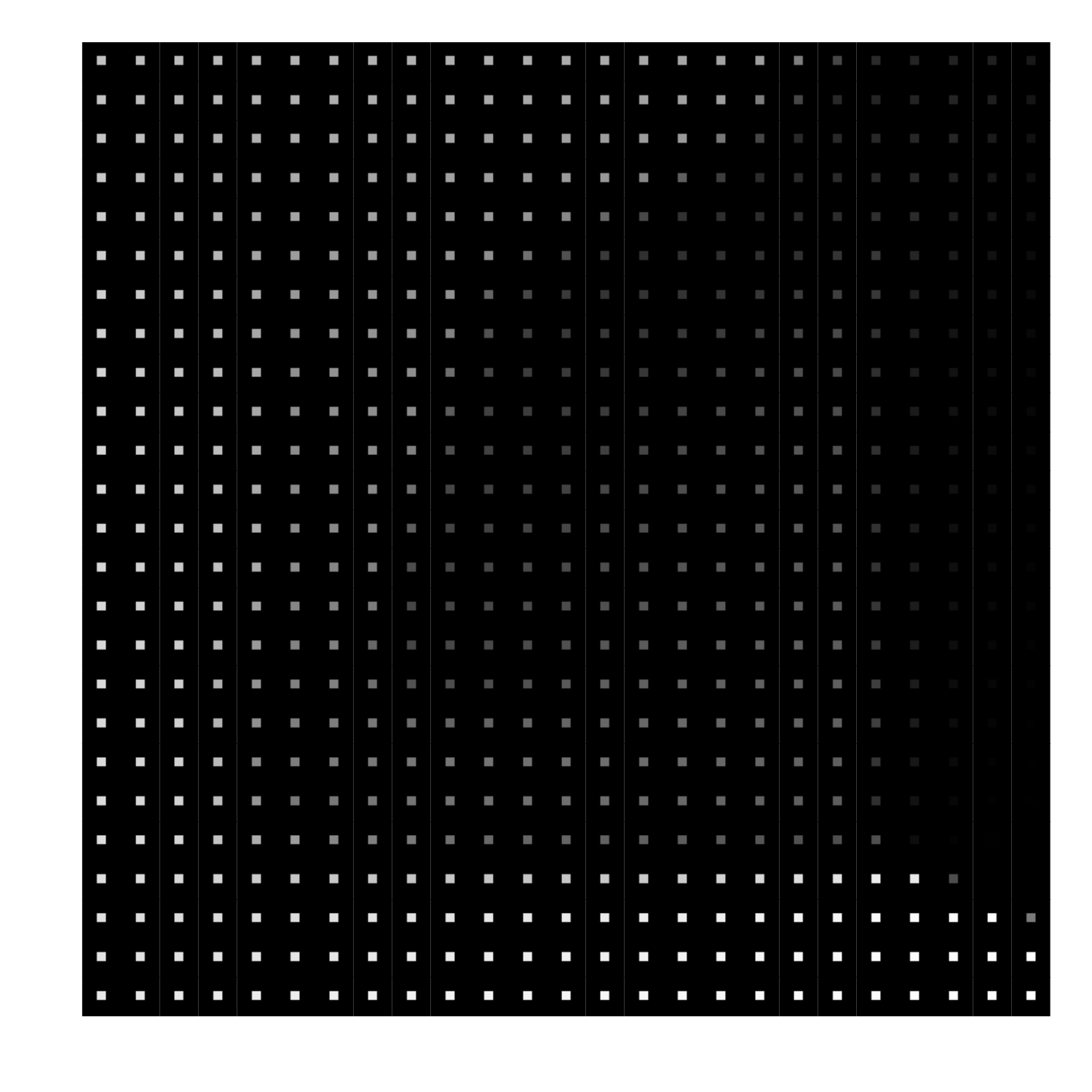}

		\includegraphics[width=\textwidth]{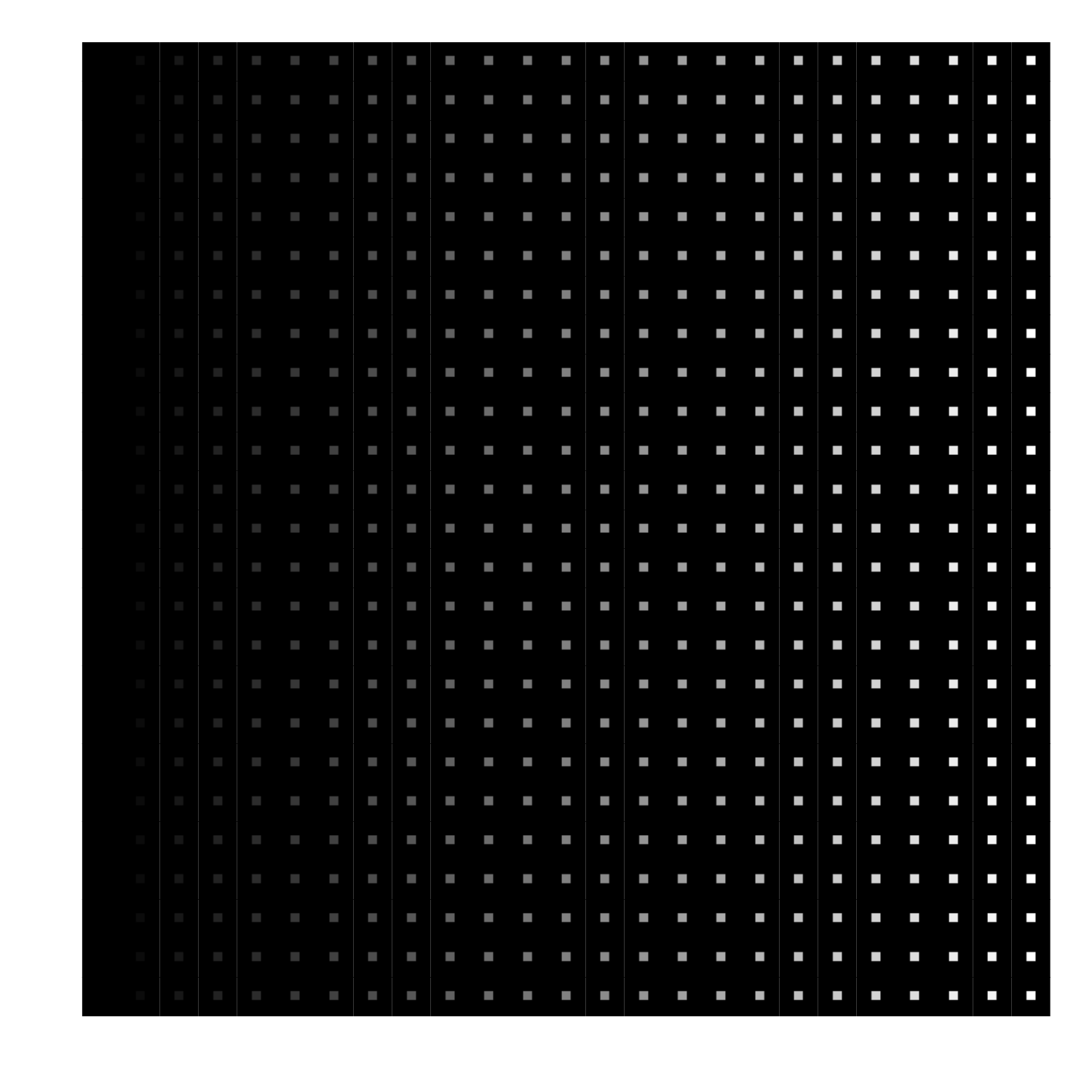}
		
	\end{minipage}%
	\caption{\label{fig:fading-squares-latent}
		Illustrations of the 2-dimensional latent space of the WAE trained on the fading squares dataset with deterministic ({\bf first row}) and random ({\bf second row}) encoders and a uniform prior $P_Z$ over the box;
		(a) 100 points sampled from the aggregated posterior $Q_Z$ ({\bf dark red}) and prior $P_Z$ ({\bf blue}); 
		(b)~same plot but with 1000 points. For the random encoder {\bf black} points show data points $x$ mapped to the mean values of the encoder $\E[Q(Z|X=x)]$;
		(c) decoder outputs at the corresponding points of the latent space.}
\end{figure*}

In this section we consider the behaviour of WAEs with deterministic and random encoders in the case of a mismatch\footnote{
It may be possible that $P_Z$ is supported on a low-dimensional subspace of $\Z$, e.g.\:on a hypersphere. 
In this case we actually mean a mismatch between $\dI$ and the dimensionality \emph{of the support of} $P_Z$.
However, this is irrelevant for the priors we consider in this work, which are either Gaussian or uniform on $[-1,1]^{\dZ}$.
} between the dimensionality of the latent space $\dZ$ and the \emph{intrinsic dimensionality} of the data distribution $\dI$, which is informally the minimum number of parameters required to continuously parametrise the data manifold.

\subsection{\label{section:fading-squares}Dimension mismatch is harmful}
What happens if a deterministic-encoder WAE is trained with a latent space of dimension $\dZ$ that is larger than the intrinsic dimensionality $\dI$? If the encoder is continuous then the data distribution $P_X$ will be mapped to $Q_Z$ supported on a latent manifold of dimension at most $\dI < \dZ$ while the regularizer in \eqref{eq:WAEobj} will encourage the encoder to fill the latent space similarly to the prior $P_Z$ as much as possible.
This is a hard task for the encoder for the same reason that it is hard to fill the plane with a one dimensional curve.

To empirically investigate this setting, we introduce the simple synthetic \emph{fading squares} dataset consisting of $32 \times 32$ pixel images of centred, $6\times 6$ pixel grey squares on a black background. The value of this colour varies uniformly from $0$ (black) to $1$ (white) in steps of $10^{-3}$. The intrinsic dimensionality of this dataset is therefore $1$, as each image in the dataset can be uniquely identified by the value of the colour of its grey square. 
 
We trained a deterministic-encoder WAE with a uniform prior over a box in a $2$-dimensional latent space on this dataset.
Since $\dZ=2$, we can easily visualise the learned embedding of the data into the latent space and the output of the decoder across the whole latent space. This is displayed in Figure \ref{fig:fading-squares-latent} (upper middle) for one such WAE.

The results of this simple experiment are illuminating about the behaviour of deterministic-encoder WAEs. The WAE is forced to reconstruct the images well, while at the same time trying to fill the latent space uniformly with the 1-dimensional data manifold. The only way to do this is by curling the manifold up in the latent space. In practice, the WAE must only fill the space to the extent that it fools the divergence measure used. The upper-left plot of Figure \ref{fig:fading-squares-latent} shows that when only a mini-batch of $100$ samples is used, it is much less obvious that the aggregate posterior does not match the prior than when we visualise the entire data manifold as in the upper-middle figure.
We found that larger mini-batches resulted in tighter curling of the manifold supporting $Q_Z$, suggesting that mini-batch size may strongly affect the performance of WAEs.

We repeated the same experiment with a random-encoder WAE, for which the encoder maps an input image to a uniform distribution over the axis aligned box with centre $(\varphi_{1}(x),  \varphi_{2}(x))$ and side lengths $(\sigma_{1}(x), \sigma_{2}(x))$. 
The lower-middle plot of Figure \ref{fig:fading-squares-latent} shows the resulting behaviour of the learned encoder. In contrast to the deterministic-encoder WAE, the random-encoder WAE is robust to the fact that $\dZ > \dP$ and can use one dimension to encode useful information to the decoder while filling the other with noise. That is, a single image gets mapped to a thin and tall box in the latent space. In this way, the random-encoder WAE is able to properly match the aggregated posterior $Q_Z$ to the prior distribution $P_Z$.

To what extent is it actually a problem that the deterministic WAE represents the data as a curved manifold in the latent space? There are two issues.

\paragraph{Poor sample quality:}
Only a small fraction of the total volume of the latent space is covered by the deterministic encoder.
Hence the decoder is only trained on this small fraction, because under the objective \eqref{eq:WAEobj} the decoder learns to act only on the encoded training images. 
While it appears in this 2-dimensional toy example that the quality of decoder samples is nonetheless good everywhere, in high dimensions, such ``holes'' may be significantly more problematic. This is a possible explanation for the results presented in Table \ref{table:dimension-fid-test-reconstruction} of Section \ref{subsection:celebA}, in which we find that large latent dimensions decrease the quality of the samples produced by deterministic WAEs\footnote{
Preliminary experimentation suggests that the better quality of samples from the {WAE-GAN} compared to the {WAE-MMD} reported in \cite{TBG+17} could be a result of the instability of the associated adversarial training. We found that when training a deterministic-encoder {WAE-GAN} on the \emph{fading squares} dataset, the {1-D} embedding of the data-manifold (the support of $Q_Z$) would move constantly through the support of $P_Z$ throughout training without converging. This means that the decoder is trained on a much larger fraction of the total volume of $P_Z$ compared to the {WAE-MMD}, for which the stability of training means that convergence to the final manifold (constituting a small fraction of $P_Z$) is quick.
}. 

\paragraph{Wrong proportion of generated images}
We found that although in this simple example all of the samples generated by the deterministic-encoder WAE are of good quality, they are not produced in the correct proportions. 
By analogy, this would correspond to a model trained on MNIST producing too few 3s and too many 7s.

To see this, consider the mean pixel value of an image in our toy dataset. It is a 1-dimensional random variable, uniformly distributed on the interval $[0, 36/1024]$, where $36/1024$ is the mean (over the whole image) in the case of a white square. We trained $5$ deterministic- and random-encoder WAEs, and for each one generated 100,000 images by sampling from the prior distribution and pushing through the learned decoder. As a baseline, we also ran this procedure using $5$ VAEs with the same architecture. We then calculated the cumulative distribution function (CDF) of the mean pixel values and compared to the theoretical uniform distribution. 

\begin{figure}[t!]
	\centering
	\includegraphics[width=\columnwidth]{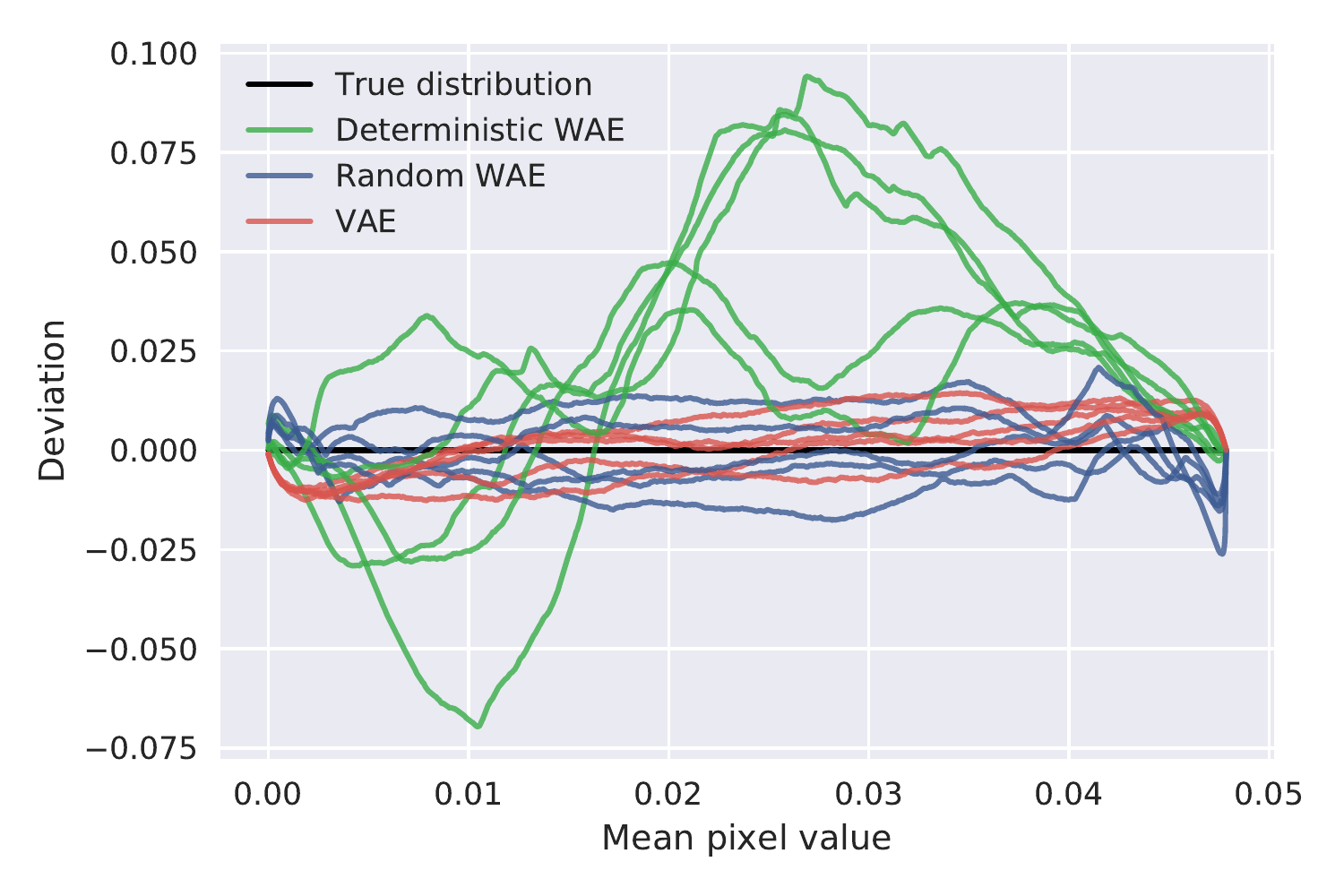}
	\caption{\label{fig:fading-squares-deviation-from-cumulative} Deviation from the correct cumulative distribution of the mean pixel values. If images were generated using the correct frequencies, the deviations should be close to $0$. The deterministic WAE does not meet this goal.}
\end{figure}
This is summarised in Figure \ref{fig:fading-squares-deviation-from-cumulative}, which displays the deviation from the theoretical CDF for each of the models trained. This shows that the deviation from the theoretical distribution for deterministic-encoder WAEs is consistently worse than for the random-encoder WAEs and VAEs, which fair comparably with one another. Note that while observing a uniform distribution here does not prove that images are generated with the correct frequencies, deviation from it does indeed indicate failure.

\subsection{\label{subsection:celebA}Random encoders with large $\dZ$}

To test our new intuitions about the behaviour of deterministic- and random-encoder WAEs with different latent dimensions, we next consider the \emph{CelebA} dataset. 
All experiments reported in this section used Gaussian priors and, for the random-encoder WAEs, Gaussian encoders. 
A~fixed convolutional architecture with cross-entropy reconstruction loss was used for all experiments. To keep computation time feasible, we used small networks.

Table \ref{table:dimension-fid-test-reconstruction} shows the results of training 5 random- and 5 deterministic-encoder WAEs with $\dZ = 32, 64, 128$ and $256$. 
We found that both deterministic- and random-encoder WAEs exhibit very similar behaviour: test reconstruction error decreases as $\dZ$ increases, while the FID scores \cite{heusel2017gans} of random samples generated by the models after training first decrease to some minimum and then subsequently increase (lower FID scores mean better sample quality).

\begin{table}[t!]
	\caption{FID scores and test reconstructions for deterministic- and random-encoder WAEs trained on \emph{CelebA} for various latent dimensions $\dZ$. Test reconstructions get better with increased dimension, while FID scores suffer for $\dZ \gg \dI$.}
	\label{table:dimension-fid-test-reconstruction}
	\vskip 0.15in
	\begin{center}
		\begin{small}
			\begin{sc}
				\resizebox{\columnwidth}{!}{%
					\begin{tabular}{lcccccr}
						\toprule
						$\dZ$ & \multicolumn{2}{c}{FID score} & \multicolumn{2}{c}{Test reconstruction (log)} \\
						&  Det. & Rand.  & Det. & Rand.\\
						\midrule
						32    & 75.0$\pm$ 0.7& 74.8$\pm$ 0.5& 6457.0$\pm$ 10.4& 6445.5$\pm$ 7.5 \\
						64 & 71.6 $\pm$ 0.8 & 71.1 $\pm$ 1.0 & 6364.4$\pm$ 7.4 & 6365.0$\pm$ 5.4 \\
						128    & 76.8$\pm$ 1.3& 76.8$\pm$ 1.2& 6300.5$\pm$ 6.6& 6309.3$\pm$ 9.7 \\
						256    & 147.6$\pm$ 2.3& 139.8$\pm$ 4.2&6265.3$\pm$ 9.5& 6262.6$\pm$ 6.7    \\
						\bottomrule
				\end{tabular}}
			\end{sc}
		\end{small}
	\end{center}
	\vskip -0.1in
\end{table}

For deterministic encoders, this agrees with the intuition we gained from the \emph{fading squares} experiment. Unable to fill the whole latent space when $\dP < \dZ$, the encoder leaves large holes in the latent space on which the decoder is never trained. When $\dP \ll \dZ$, these holes occupy most of the total volume, and thus most of the samples produced by the decoder from draws of the prior are poor.

For random encoders we did not expect this behaviour. Rather than automatically filling unnecessary dimensions with noise when $\dP\ll\dZ$ similarly to the fading squares example, thus making $Q_Z$ accurately match $P_Z$ and preserving good sample quality, the random encoders would ``collapse'' to deterministic encoders. That is, the variances of $Q(Z|X=x)$ tend to $0$ for almost all dimensions and inputs $x$.\footnote{More specifically, we parametrised the \emph{log-variances} for each dimension and observed that the maximum of these on any mini-batch of data would typically be smaller than $-10$.} The fact that this happens for most---but not all---dimensions explains why with a $256$ dimensional latent space, random WAEs produced samples with bad FID scores, but still better than deterministic WAEs: to some extent dimesionality reduction \emph{does} occur, just not as much as we would hope for.

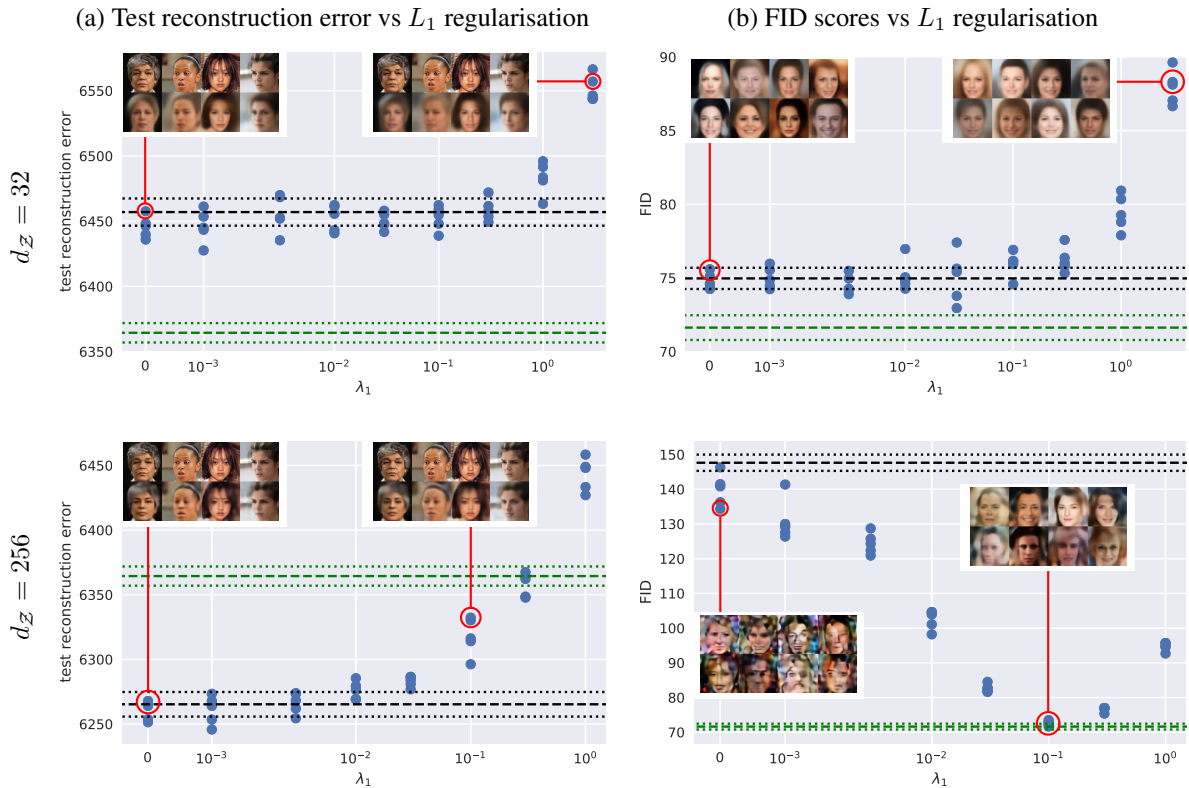
\begin{figure*}[t!]
	\centering
	\begin{minipage}{.02\textwidth}
		\begin{turn}{90}
			\hspace{1cm} $\dZ =256$\hspace{3.5cm} $\dZ =32$\hspace{1cm}
		\end{turn}       
	\end{minipage}
	\begin{minipage}{.45\textwidth}
		\centering
		(a) Test reconstruction error vs $L_1$ regularisation
		\begin{tikzpicture}
		\node[anchor=south west,inner sep=0] (image) at (0,0) {		\includegraphics[width=\columnwidth]{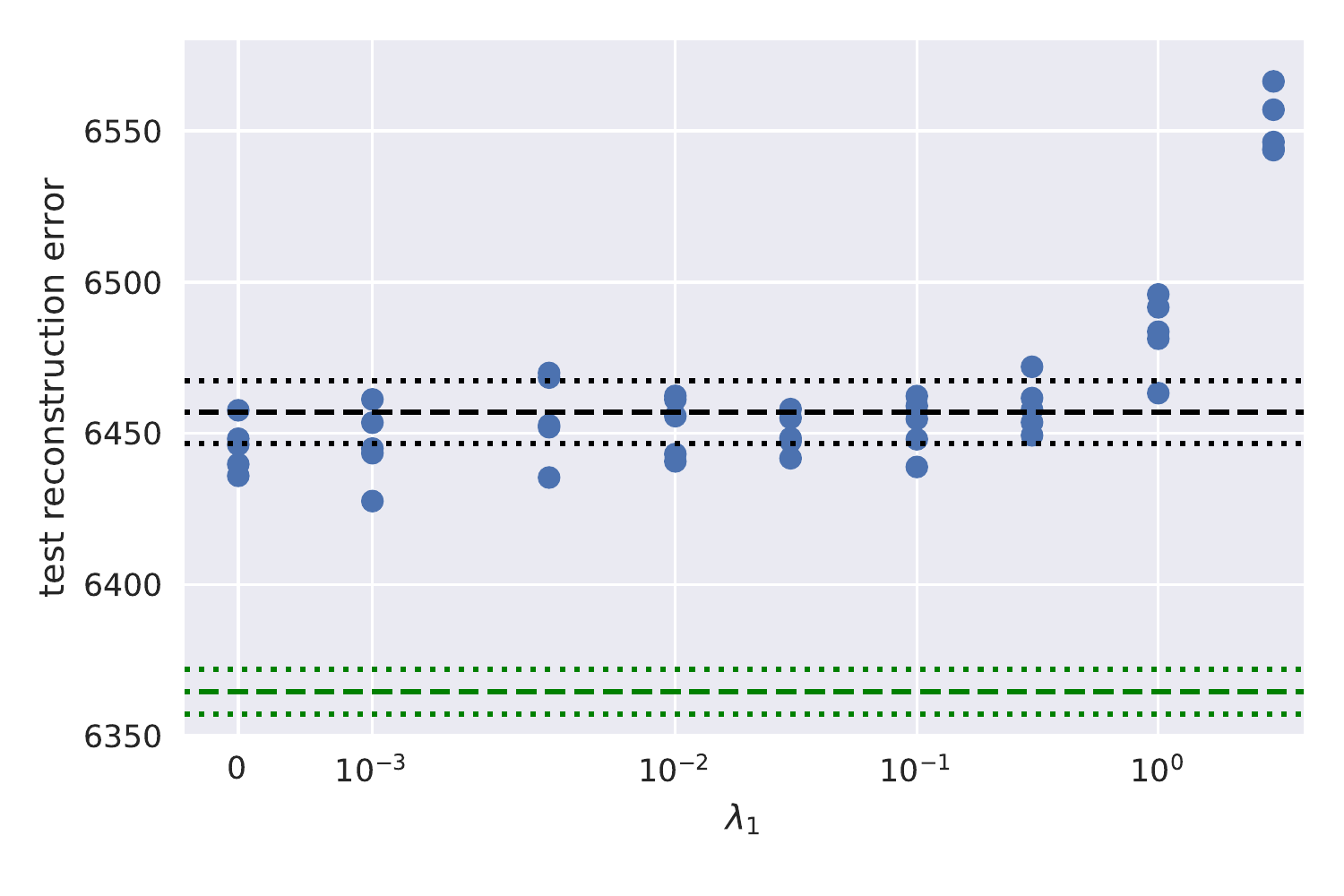}};
		\begin{scope}[x={(image.south east)},y={(image.north west)}]
		\draw[thick, red] (0.177,0.81) -- (0.177,0.56){};
		\node (recon32) at (0.27,0.845) {
			\includegraphics[width=0.3\columnwidth]{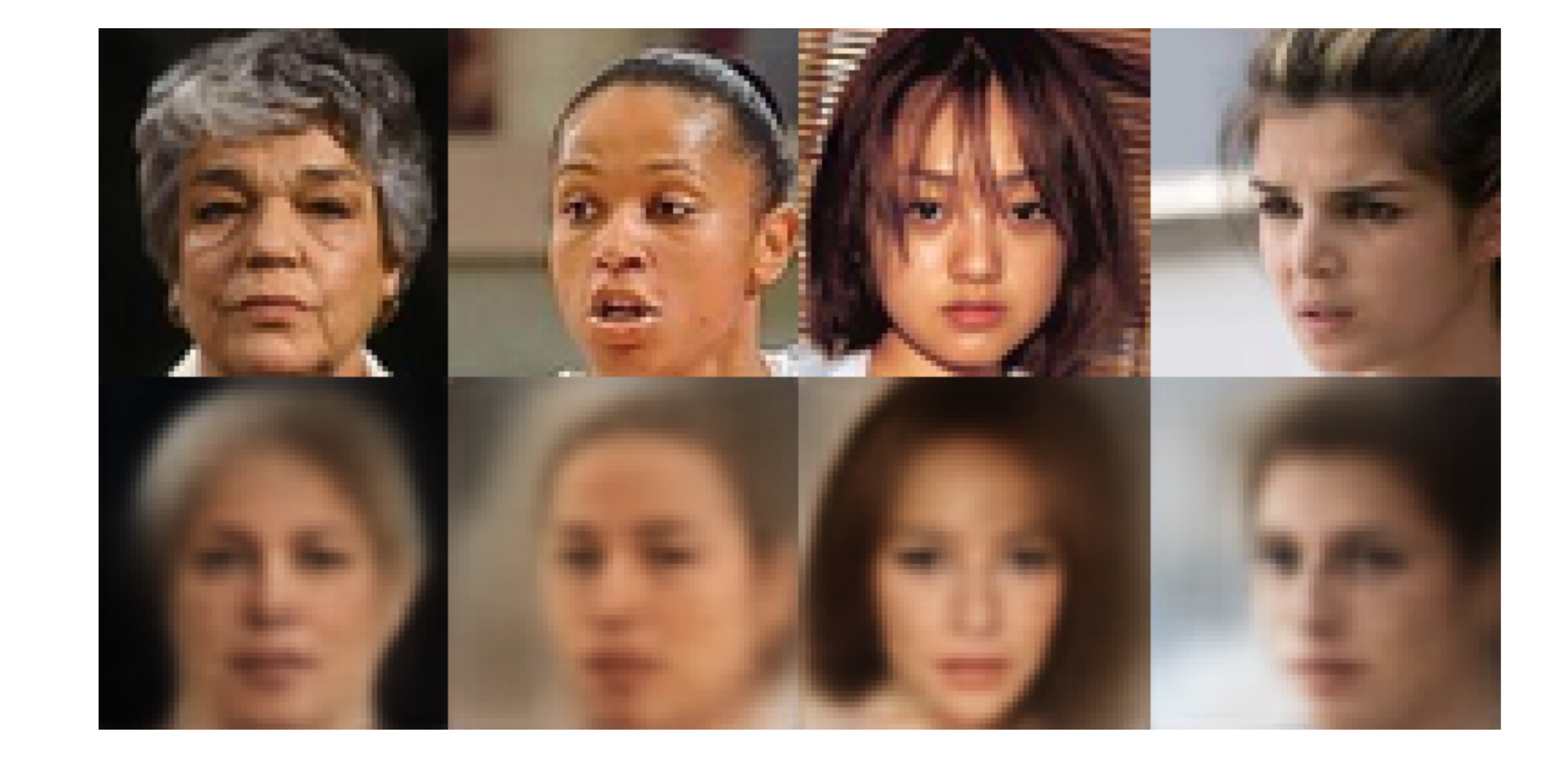}};
		\draw[red, thick,rounded corners] (0.177,0.543) circle (0.1cm);
		\draw[thick, red] (0.72,0.877) -- (0.9362,0.877){};
		\node (recon32) at (0.7,0.845) {
			\includegraphics[width=0.3\columnwidth]{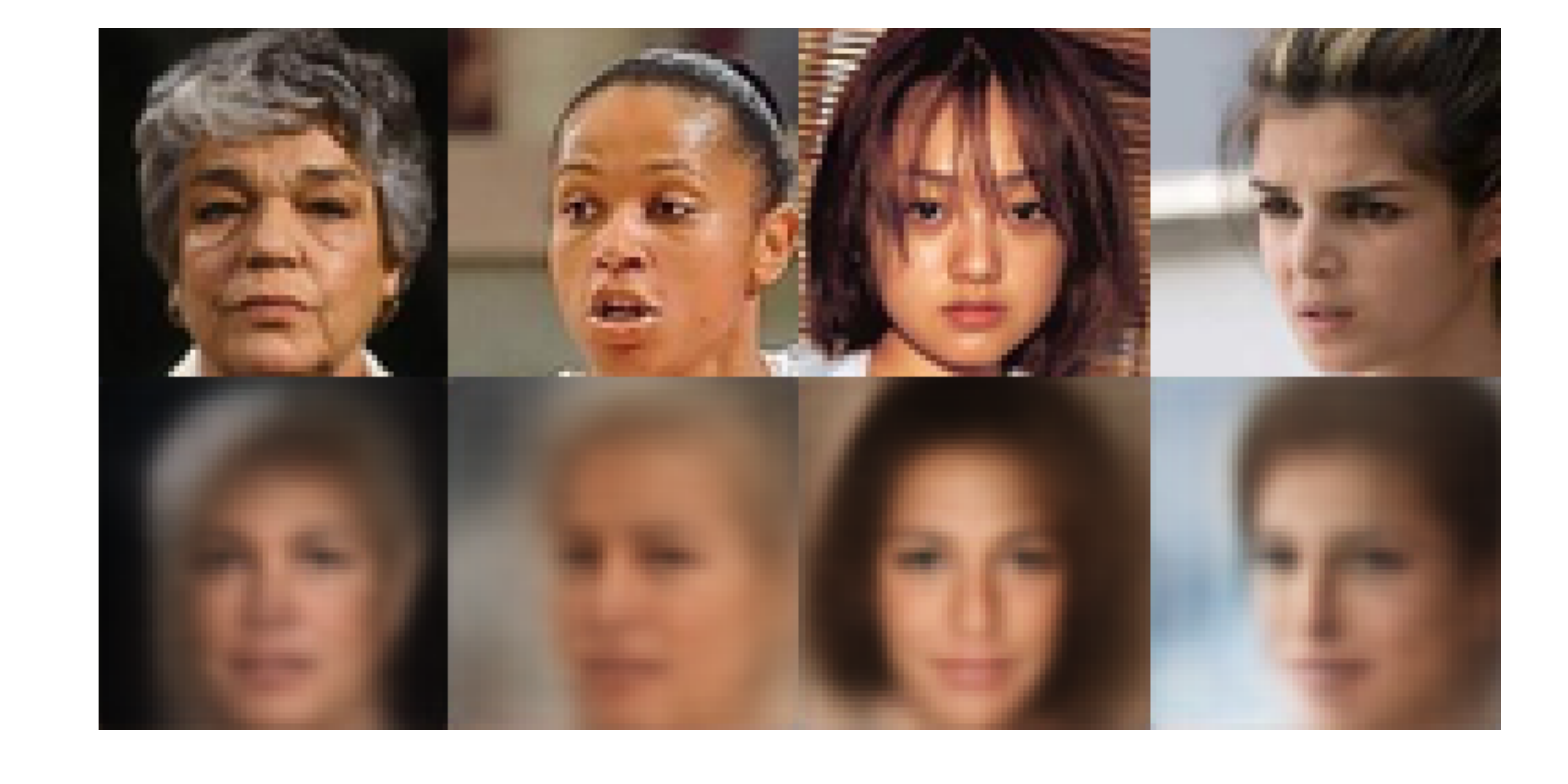}};
		\draw[red, thick,rounded corners] (0.947,0.877) circle (0.1cm);
		\end{scope}
		\end{tikzpicture}
		
		\begin{tikzpicture}
		\node[anchor=south west,inner sep=0] (image) at (0,0) {		\includegraphics[width=\columnwidth]{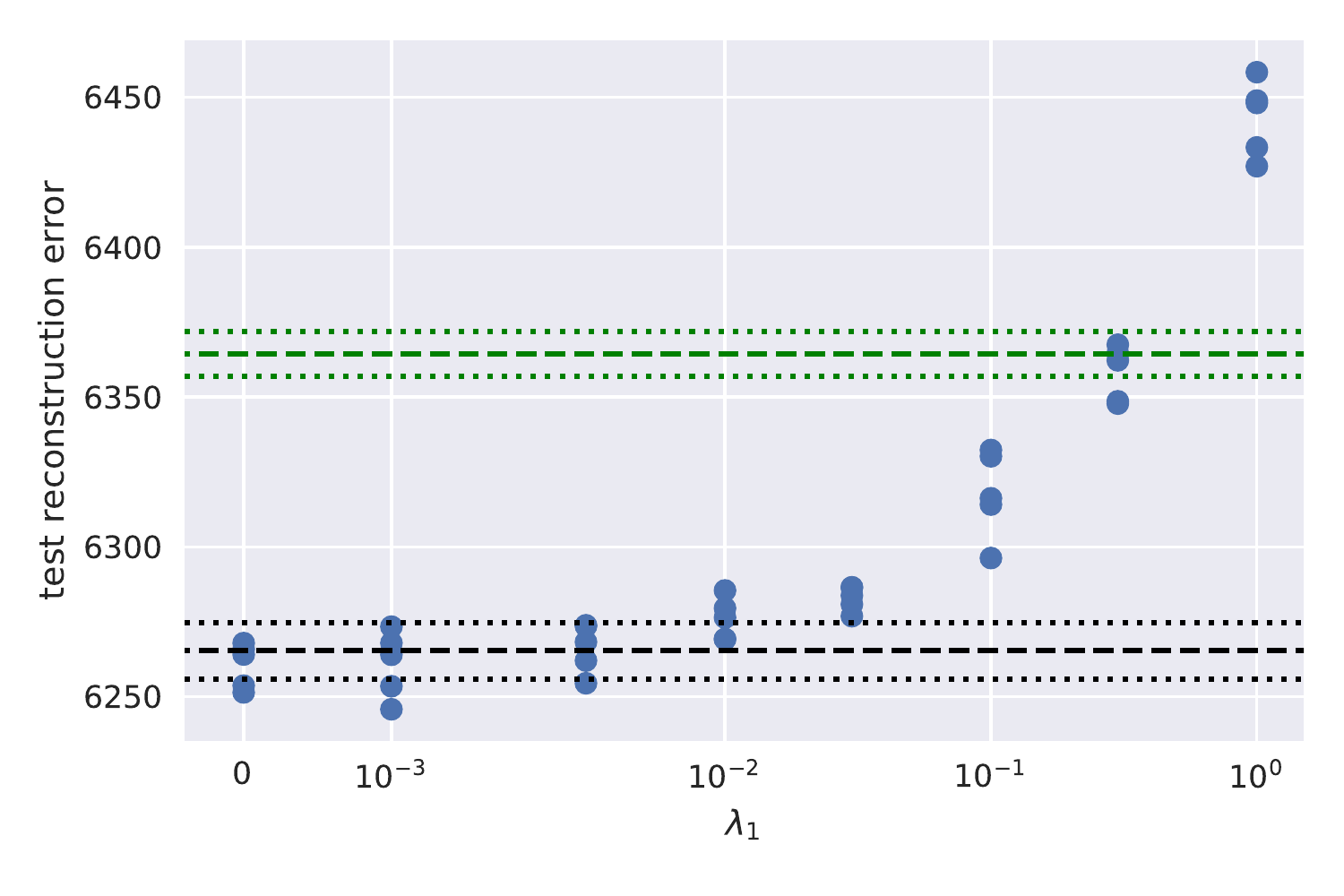}};
		\begin{scope}[x={(image.south east)},y={(image.north west)}]
		\draw[thick, red] (0.182,0.81) -- (0.182,0.31){};
		\node (recon32) at (0.27,0.845) {
			\includegraphics[width=0.3\columnwidth]{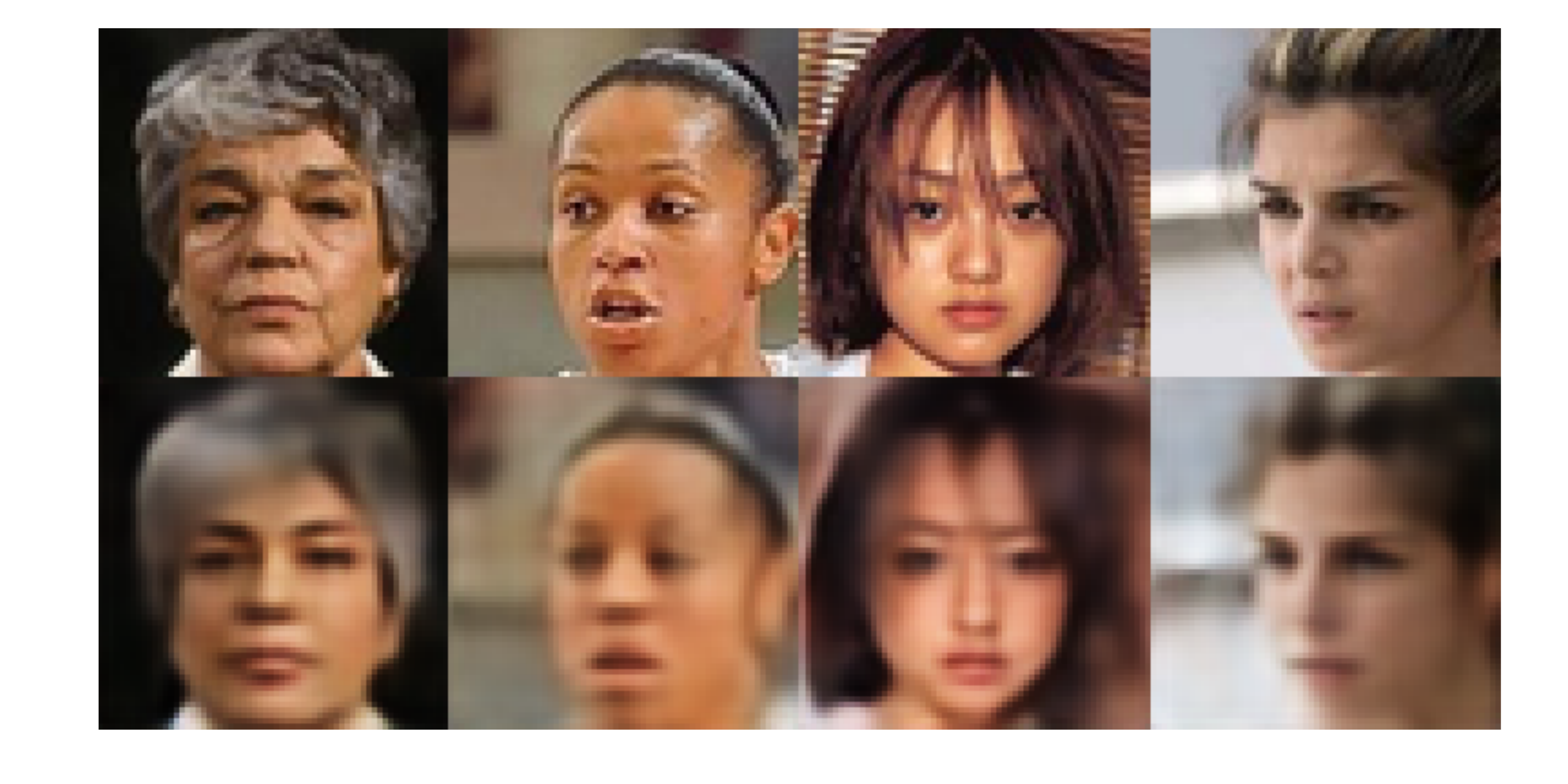}};
		\draw[red, thick,rounded corners] (0.182,0.28) circle (0.15cm);
		\draw[thick, red] (0.737,0.885) -- (0.737,0.52){};
		\node (recon32) at (0.7,0.845) {
			\includegraphics[width=0.3\columnwidth]{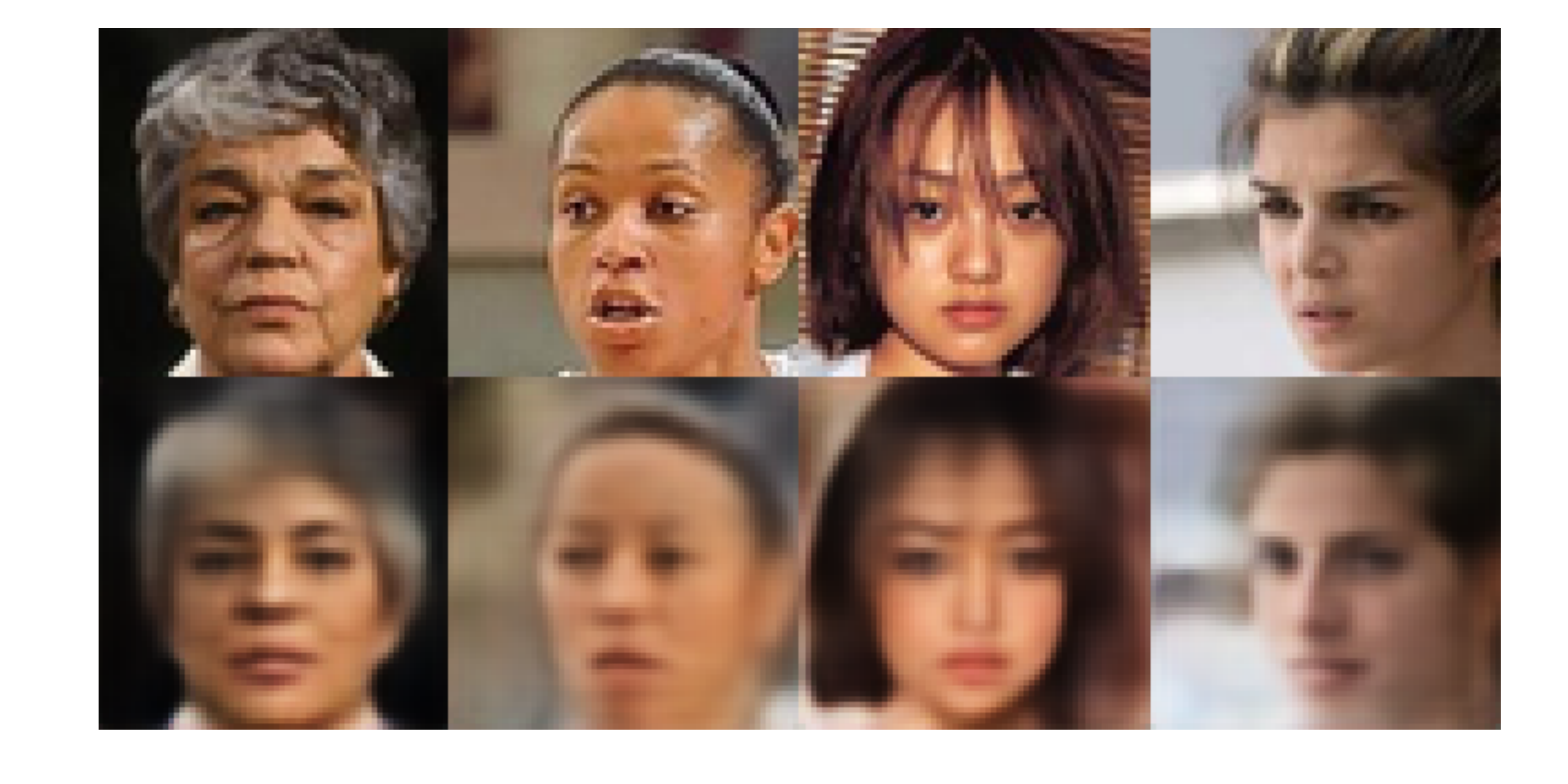}};
		\draw[red, thick,rounded corners] (0.737,0.498) circle (0.13cm);
		\end{scope}
		\end{tikzpicture}

	\end{minipage}%
	\begin{minipage}{.45\textwidth}
		\centering
		(b) FID scores vs $L_1$ regularisation
		
		\begin{tikzpicture}
		\node[anchor=south west,inner sep=0] (image) at (0,0) {		\includegraphics[width=\columnwidth]{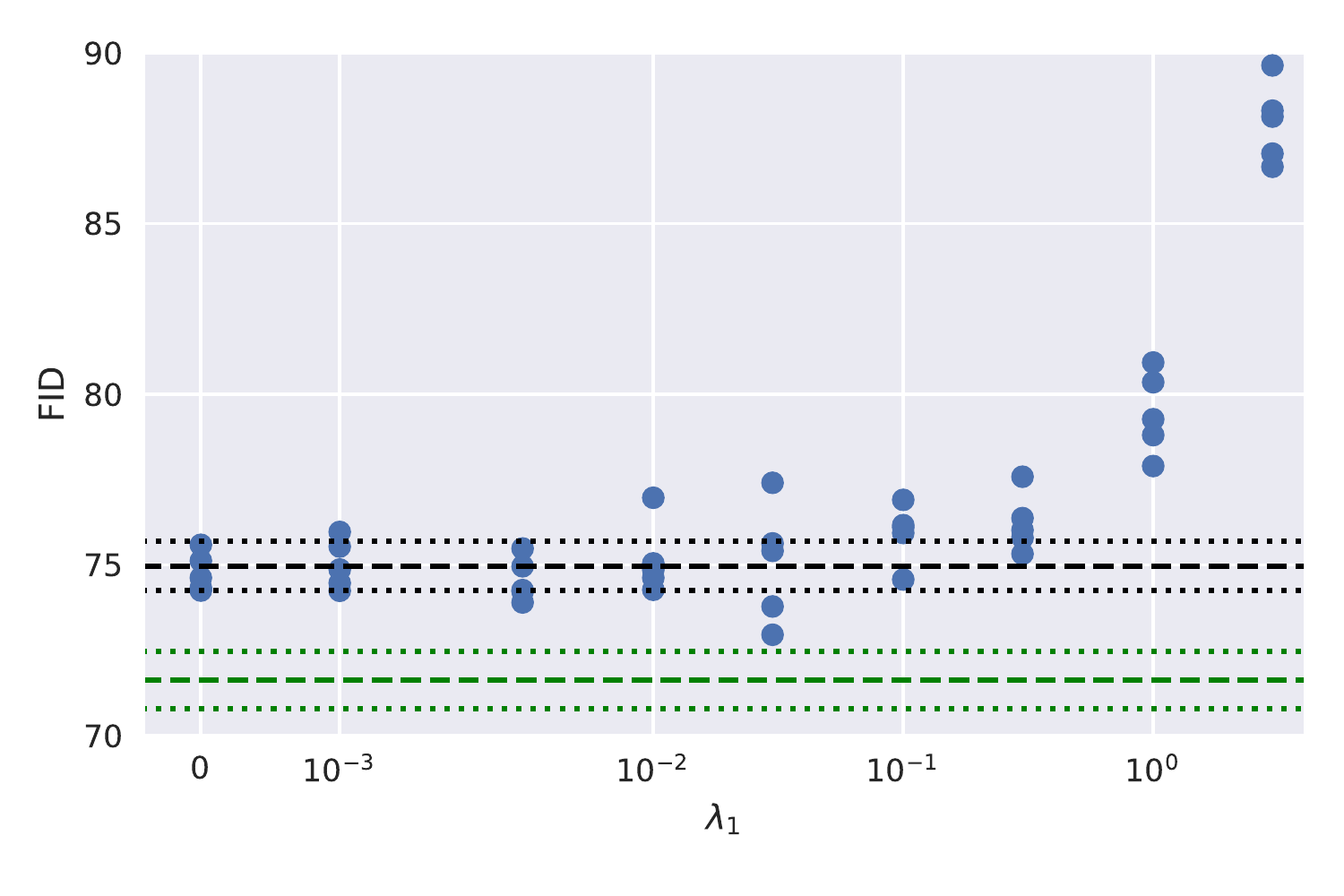}};
		\begin{scope}[x={(image.south east)},y={(image.north west)}]
		\draw[thick, red] (0.15,0.81) -- (0.15,0.415){};
		\node (recon32) at (0.25,0.83) {
			\includegraphics[width=0.3\columnwidth]{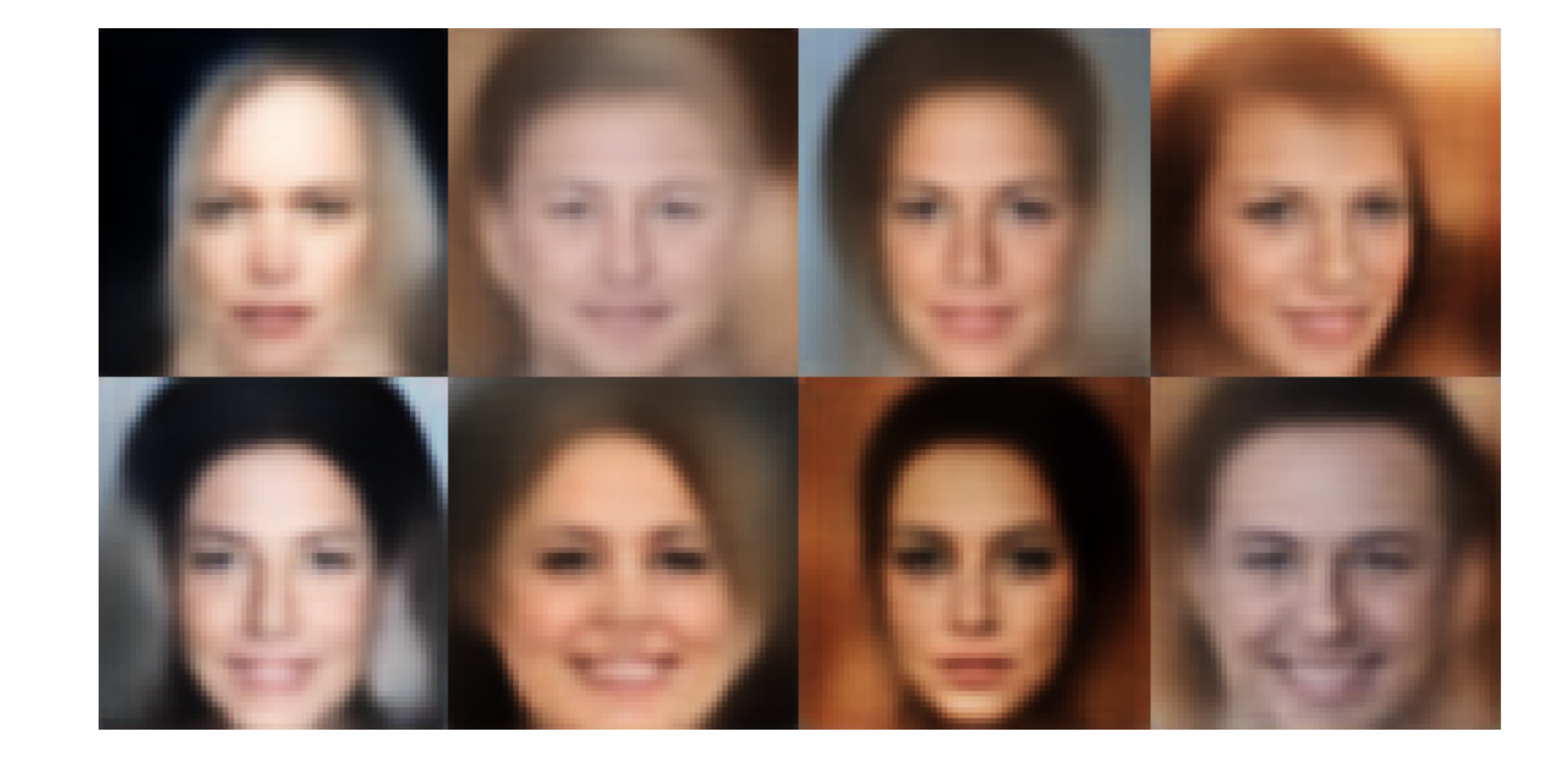}};
		\draw[red, thick,rounded corners] (0.15,0.39) circle (0.13cm);
		\draw[thick, red] (0.72,0.876) -- (0.928,0.876){};
		\node (recon32) at (0.7,0.83) {
			\includegraphics[width=0.3\columnwidth]{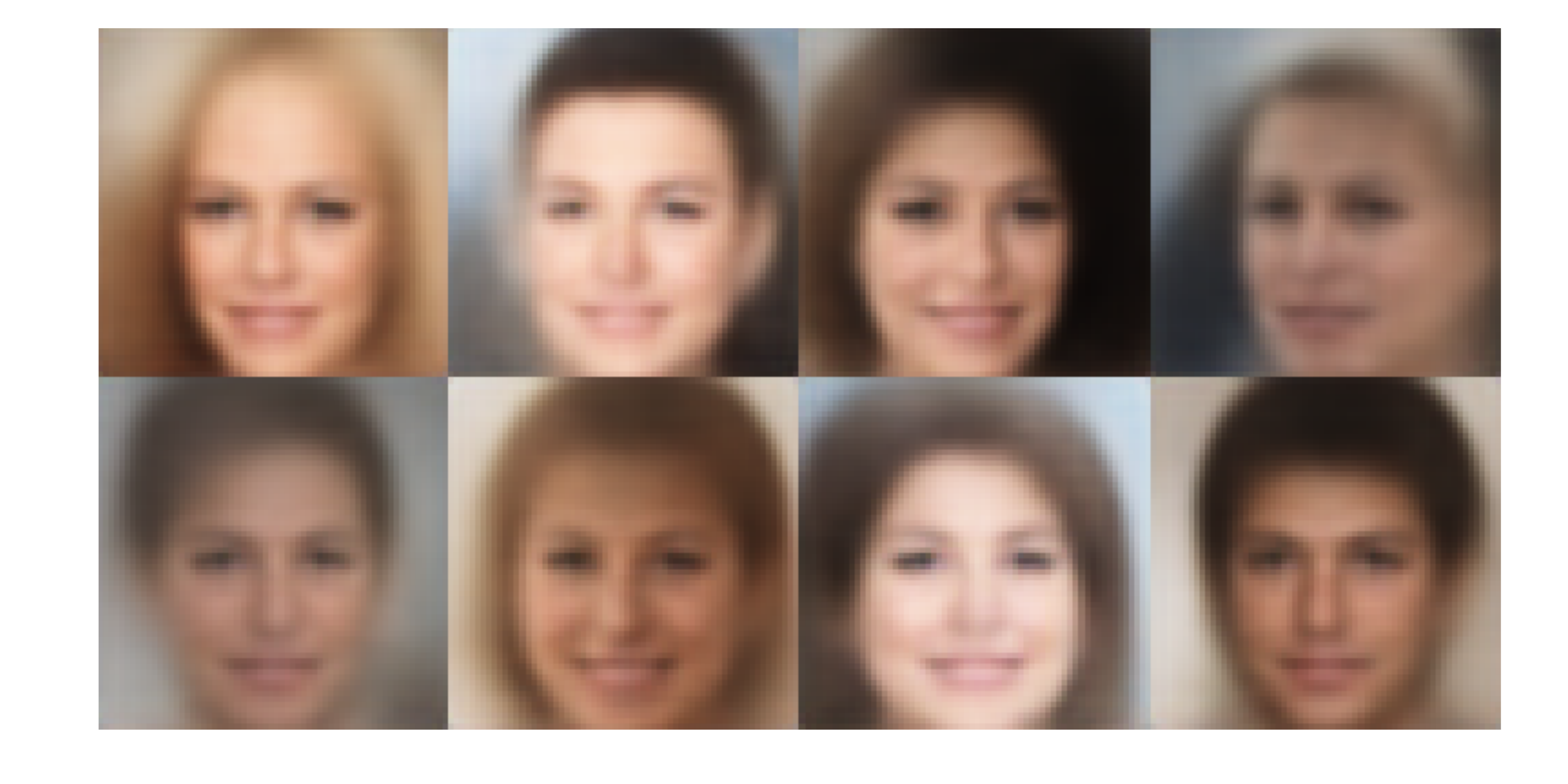}};
		\draw[red, thick,rounded corners] (0.946,0.876) circle (0.15cm);
		\end{scope}
		\end{tikzpicture}
		
		\begin{tikzpicture}
		\node[anchor=south west,inner sep=0] (image) at (0,0) {		\includegraphics[width=\columnwidth]{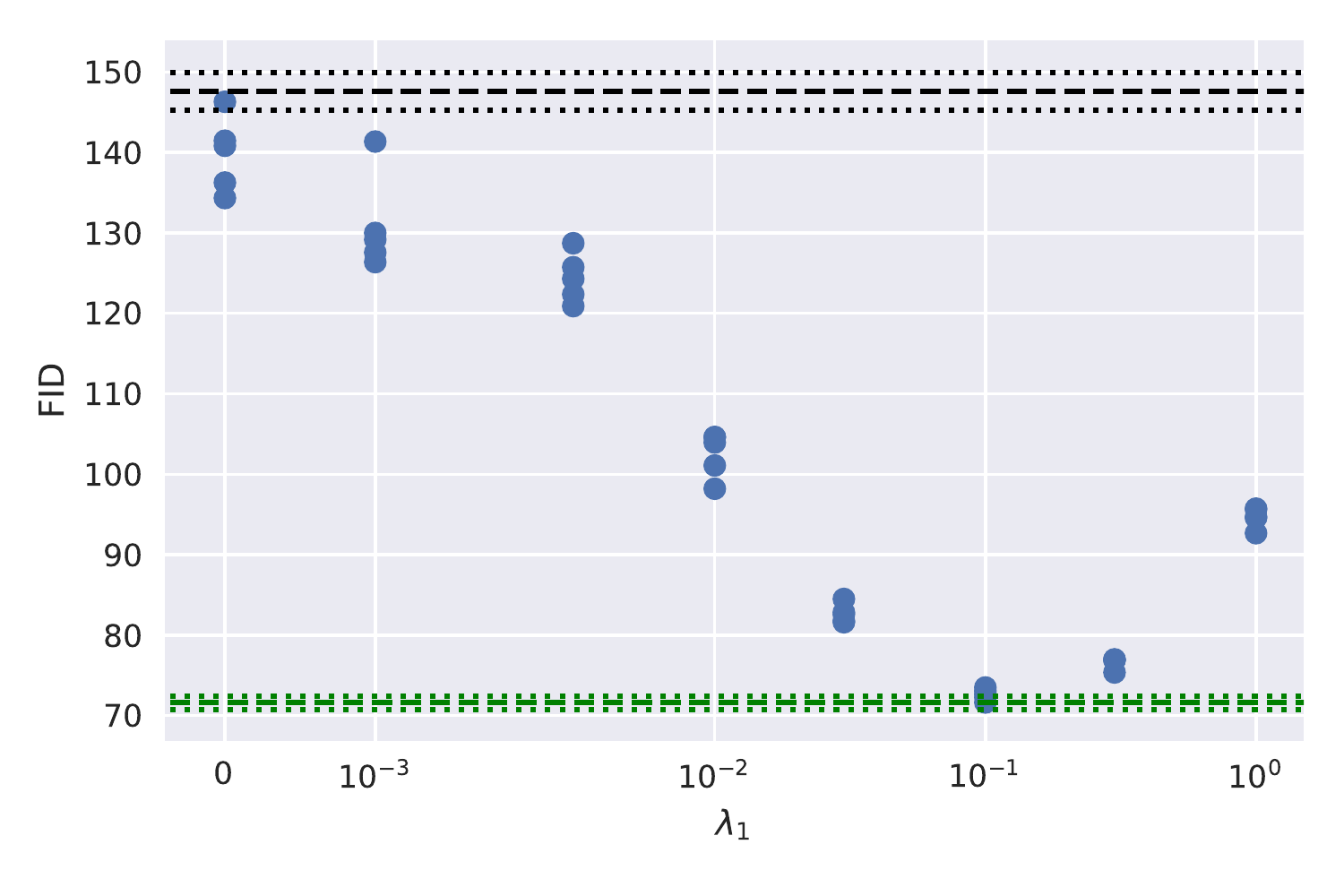}};
		\begin{scope}[x={(image.south east)},y={(image.north west)}]
		\draw[thick, red] (0.168,0.758) -- (0.168,0.415){};
		\node (recon32) at (0.265,0.4) {
			\includegraphics[width=0.3\columnwidth]{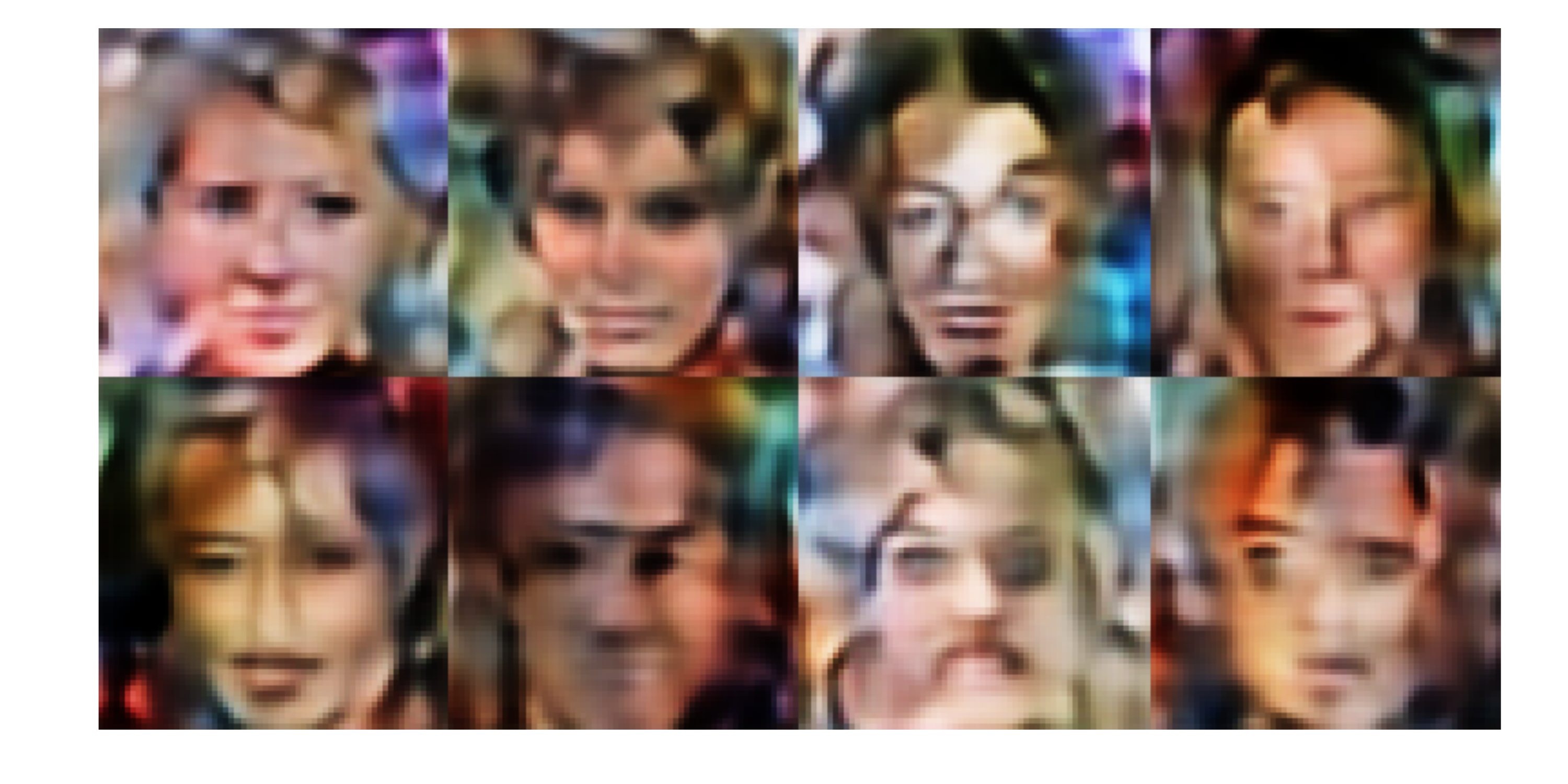}};
		\draw[red, thick,rounded corners] (0.168,0.78) circle (0.1cm);
		\draw[thick, red] (0.732,0.73) -- (0.732,0.255){};
		\node (recon32) at (0.73,0.73) {
			\includegraphics[width=0.3\columnwidth]{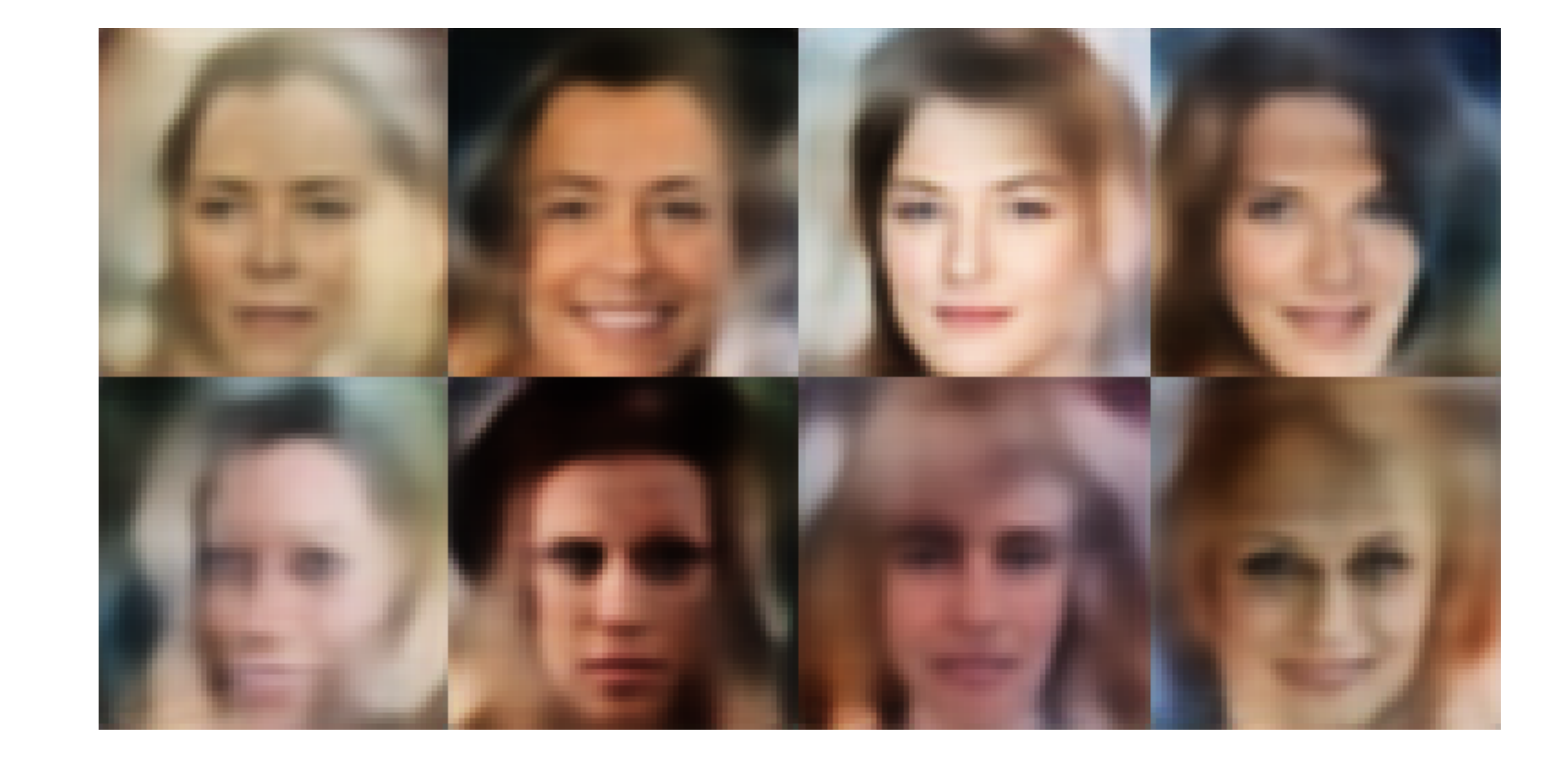}};
		\draw[red, thick,rounded corners] (0.732,0.225) circle (0.15cm);
		\end{scope}
		\end{tikzpicture}

	\end{minipage}%
	\caption{\label{fid:celebA-fid-test-errors-vs-latent-dim}FID scores and test reconstruction errors for random-encoder WAEs with latent space dimension $\dZ=32$ (\textbf{first row}) and $\dZ=256$ (\textbf{second row}) for different $L_1$ regularisation coefficients $\lambda_1$. In each plot, the dashed/dotted black lines represent the mean $\pm$ s.d. for deterministic-encoder WAEs with the same $\dZ$ (i.e. $32$ or $256$). The dashed/dotted green lines represent the mean $\pm$ s.d. for deterministic WAEs $\dZ=64$, for which the FID scores were best amongst all latent dimensions we tested. Overlaid images are (a) test reconstructions and (b) random samples coming from experiments indicated by the red circle. These plots show that when $\dZ<\dP$, (i) random-encoder WAEs perform comparably to deterministic WAEs and (ii) when appropriately regularised ($\lambda=10^{-1}$), random encoders with high dimensional latent spaces are able to produce samples of similar quality to the deterministic encoders with the best latent space dimension. At the same time, the test reconstruction errors are lower.}
\end{figure*}

\paragraph{Resolving variance collapse through regularization}
The cause of this variance collapse is uncertain to us. Two possible explanations are a problem of optimization or a failing of the MMD as a divergence measure. 
We found, however, that we could effectively eliminate this issue by adding additional regularisation in the form of an $L_p$ penalty on the log-variances. This provides encouragement for the variances to remain closer to $1$ and thus for the encoder to be remain stochastic. More precisely, we added the following term to the objective function to be minimised:
\begin{equation}
\frac{\lambda_{p}}{N}\sum_{n=1}^N \sum_{i=1}^{\dZ} \left|\log\left(\sigma^2_{i}(x_n)\right) \right|^p
\end{equation}
where $i$ indexes the dimensions of the latent space $\Z$, $n$ indexes the inputs in a mini-batch and $\lambda_{p}\geq 0$ is a new regularization coefficient.

We experimented with both $L_1$ and $L_2$ regularisation. We found similar qualitative behaviour with both, but found $L_1$ regularisation to give better performance and thus report only these results. Note that an $L_1$ penalty on the log-variances should encourage the encoder/decoder pair to \emph{sparsely} use latent dimensions to code useful information. Indeed, the $L_1$ penalty will encourage sparsely many dimensions to have non-zero log-variances, and if the variance of $Q(Z|X)$ in some dimension $i$ is always $1$ then in order for the marginal $Q(Z_i)$ to match $P(Z_i)$, we must have that $\varphi_i(x)=0$ for all inputs $x$. 

Using latent dimensions 32 and 256 to consider both the case of under- and over-shooting the intrinsic dimensionality of the dataset,\footnote{The fact that the deterministic WAE produced samples with better FIDs with $\dZ=64$ than $\dZ=32$ suggested to us that the intrinsic dimensionality of the CelebA datset is greater than 32.} we trained 5 $L_1$-regularized random-encoder WAEs for a variety of values for $\lambda_1$. Figure \ref{fid:celebA-fid-test-errors-vs-latent-dim} shows the test reconstruction errors and FID scores obtained at the end of training.

When $\dI \ll \dZ$, $L_1$ regularisation can significantly improve the performance of random-encoder WAEs compared to their deterministic counterparts. In particular, tuning for the best $\lambda_1$ parameter results in samples of quality comparable to deterministic encoders with the best latent dimension size, while simultaneously achieving lower test reconstruction errors.
Through appropriate regularisation, random-encoder WAEs are able to adapt to the case that $\dZ \gg \dP$ and still perform well.

When $\dI > \dZ$,  $L_1$ regularisation does not improve test reconstruction error and FID scores and the random-encoder WAEs perform at best the same as deterministic-encoder WAEs. This makes sense: if in the deterministic case the WAE is already having to perform ``lossy compression'' by reducing the effective dimensionality of the dataset, then the optimal random encoder cannot do better than becoming deterministic. Thus, forcing the encoder to be more random can only harm performance.

The reader will notice that we have merely substituted the problem of searching for the ``right'' latent dimensionality $\dZ$ with the problem of searching for the ``right'' regularisation $\lambda_1$. 
However, these results show that random encoders are capable of adapting to the intrinsic data dimensionality; future directions of research include exploring divergence measures other than MMD and whether the $L_p$ regularisation coefficients $\lambda_p$ can be adaptively adjusted by the learning machine itself.

\section{\label{section:disentanglement}Learned representation and disentanglement}

\begin{figure*}[t!]
	\centering
	\begin{subfigure}[t]{0.5\textwidth}
		\centering
		\includegraphics[width=\columnwidth]{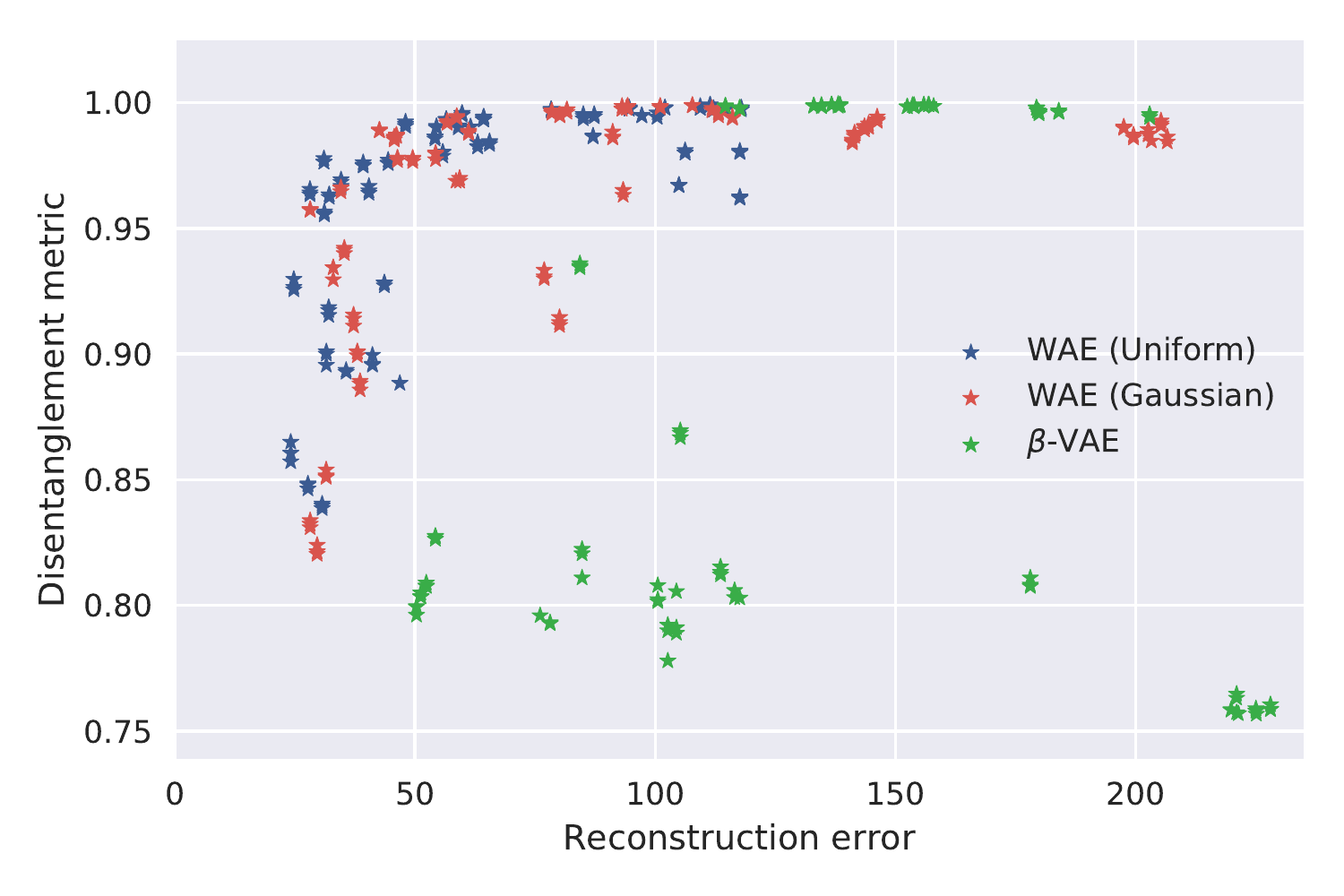}
		\caption{\label{subfig:disentanglement-vs-reconstruction-4}4-variable \emph{dSprites} disentanglement task.}
	\end{subfigure}%
	~ 
	\begin{subfigure}[t]{0.5\textwidth}
		\centering
		\includegraphics[width=\columnwidth]{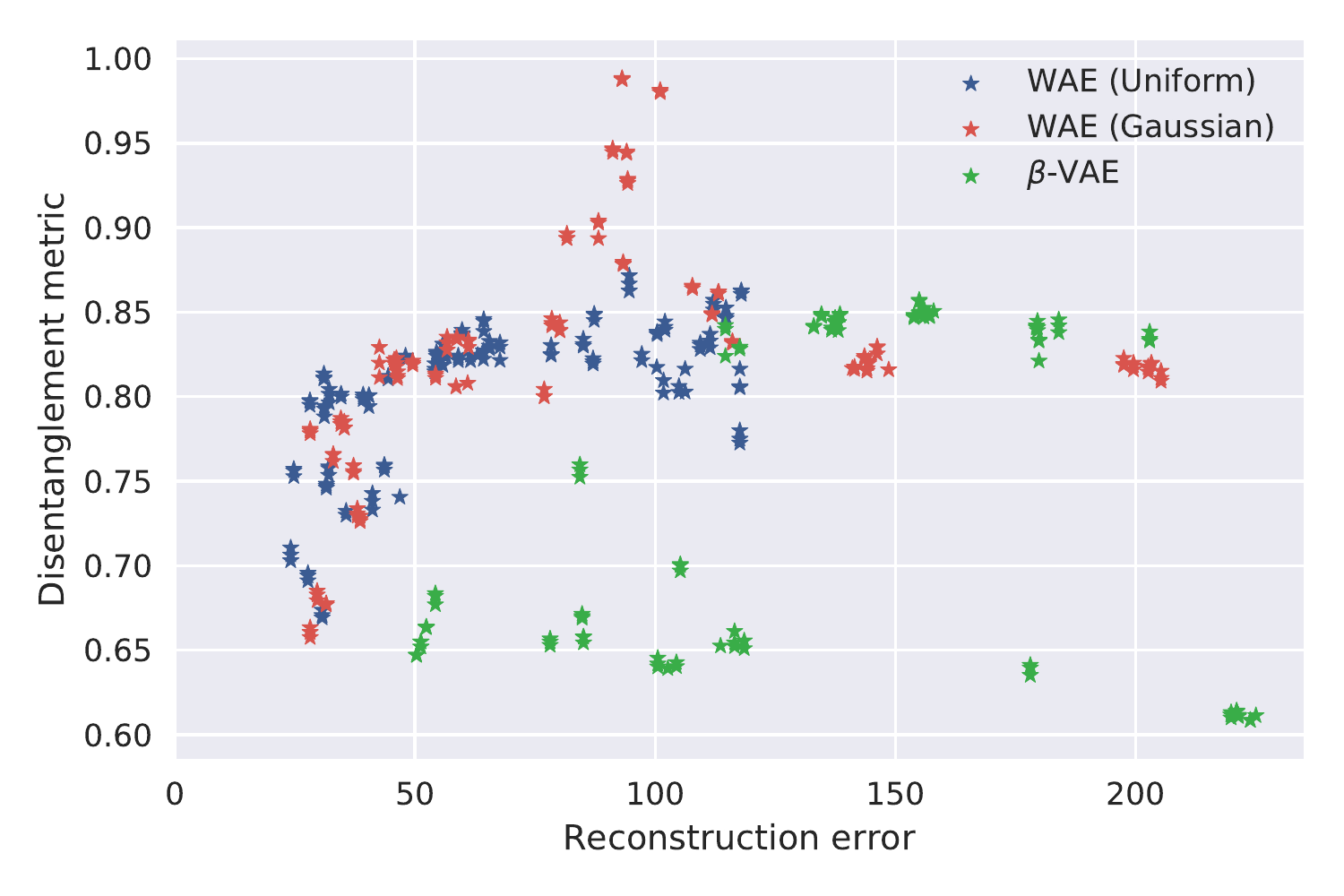}
		\caption{\label{subfig:disentanglement-vs-reconstruction-5}5 variable \emph{dSprites} disentanglement task.}
	\end{subfigure}
	\caption{\label{fig:disentanglement-vs-reconstruction}Disentanglement vs reconstruction error for $\beta$-VAEs with various values of $\beta$ and WAEs with various $L_1$ regularisation coefficients $\lambda_1$ \textbf{(up and left is better)}. Note that there is no direct way to compare different values of $\beta$ and $\lambda_1$, but in both cases increasing the value of the hyper-parameter is correlated with increasing reconstruction error. \textbf{WAEs are capable of achieving comparable or better disentanglement scores than the $\beta$-VAE while simultaneously achieving lower reconstruction errors.} In particular, WAE attains a maximum $\mathbf{98.8\%}$ on the 5-variable disentanglement tast, compared to a maximum of $\mathbf{85.4\%}$ for $\beta$-VAE)}
\end{figure*}

\emph{Disentangled representation learning} is closely related to the more general problem of \emph{manifold learning} for which auto-encoding architectures are often employed. The goal, though not precisely defined, is to learn representations of datasets such that individual coordinates in the feature space correspond to human-interpretable generative factors  (also referred to as \emph{factors of variation} in the literature). It is argued by \citet{bengio2013representation} and \citet{lake2017building} that learning such representations is essential for significant progress in machine learning research.

\subsection{A benchmark disentanglement task}

Recently, \citet{HM+17} proposed the synthetic \emph{dSprites} dataset and a metric to evaluate algorithms on their ability to learn disentangled representations. The dataset consists of $2$-dimensional white shapes on a black background with $5$ factors of variation: shape, size, rotation, $x$-position and $y$-position. Samples from this dataset can be seen in the first row of Figure \ref{fig:dsprites-reconstructions}.

The metric can be used to evaluate the ``level of disentanglement'' in the representation learned by a model when the ground truth generative factors are known for each image, such as for the \emph{dSprites} dataset. We provide here an intuition of what the metric does; see \citet{HM+17} for full details. Given a trained feature map $\varphi\colon \X\to\Z$ from the image space to the latent space, we ask the following question. Suppose we are given two images $x_1$ and $x_2$ which have exactly one latent factor whose value is the same---say they are both the same shape, but different in size, position and rotation. 
 By looking at the \emph{absolute values of the difference in feature vectors} ${|\varphi(x_1) - \varphi(x_2)|}\in\R^{\dZ}$, is it possible to identify that it is the \emph{shape} that they share in common, and not any other factor?

The idea is that if a disentangled representation has indeed been learned, then for each latent factor there should be some feature coordinate $\varphi_i$ corresponding to it. The value of ${|\varphi_i(x_1) - \varphi_i(x_2)|}$ should then be close to zero for the latent factor that is shared, while other coordinates should on average be larger. 

In the same paper, the authors introduce the $\beta$-VAE, which is currently considered to be the state-of-the-art in disentangled learning algorithms. The $\beta$-VAE is a modification of the original VAE in which the KL regularisation term is multiplied by a scalar hyper-parameter $\beta$.
The authors show that by tuning $\beta$, they are able to explore a trade-off between entangled representations with low reconstruction error and disentangled representations with high reconstruction error.

We believe that WAEs have advantages over the $\beta$-VAE for disentangled representation learning, which can be considered a special case of manifold learning. The results of \citet{TBG+17} show that WAEs can learn to produce better quality samples than VAEs, suggesting that in some cases WAEs are able to learn a representation of the data-manifold better than VAEs. 

Further to this, we observe that the flexibility of the WAE framework allows arbitrary choices of prior $P_Z$ and encoder $Q(Z|X)$ distributions with only trivial changes to implementation in code, meaning that the model can be explicitly endowed with prior knowledge about the possible underlying generative factors of the dataset on which training is taking place. 

In particular, priors with different topologies can be easily used with the WAE framework. For instance, a uniform distribution over the circle $\mathcal{S}^1$ could be used to model the factor of rotation, for which rotations of $0$ and $2\pi$ should be considered the same; priors with such non-trivial topologies could be combined to encode complex knowledge, such as the presence of a circular variable (rotation), a discrete variable (shape), and three uniform variables (x-position, y-position and scale) in the dSprites dataset.

While fully investigating the possible role of using different prior distributions was outside of the scope of this project, we felt that our initial results in this direction would be of sufficient scientific interest to report here.
\vspace{-0.3cm}
\subsection{Learning disentangled representations with WAEs}

We carefully replicated the main experiment performed by \citet{HM+17} on the \emph{dSprites} dataset, which we describe in brief here. For further details, we refer the reader to Section 4.2 and Appendix A.4 of their paper.

Following the same procedure as \citet{HM+17}, we used a fixed fully connected architecture with the Bernoulli reconstruction loss for all experiments, with a latent space dimension of 16. We trained $10$ $\beta$-VAEs for $\beta\in\{1, 3, 10, 20, 30, 40, 50 ,75, 100\}$. For each of the $10$ replicates of each value of $\beta$, we calculated the disentanglement metric $3$ times. From the resulting list of $30$ numbers, we discarded the bottom 50\%. For each experiment, we also record the test reconstruction error on a held out part of the dataset. At the end of this procedure we had 15 pairs of numbers (test reconstruction error, disentanglement) for each of the 9 choices of $\beta$. 

We repeated the same process with two types of random-encoder WAEs sharing the same architectures as the $\beta$-VAE for the encoder and decoder. The first type had Gaussian priors and Gaussian encoders. The second type had a uniform prior on $[-1,1]^{\dZ}$ and uniform encoder\footnote{Here the \emph{log-side-lengths} were parametrised by the encoder, not the log-variances.} mapping to axis-aligned boxes in $\Z$.
In both cases, the means of the encoders were constrained to be in the range $(-1,1)$ on each dimension by \emph{tanh} activation functions.
We trained such WAEs with $L_1$ regularisation coefficients $\lambda_1 \in \{0, 0.1, 0.5, 1, 2, 3, 5, 8, 12\}$.

\citet{HM+17} report their results for disentangling on only 4 of the possible 5 variables.\footnote{Although there are 5 factors of variation in the \emph{dSprites} dataset, the number $99.23\pm0.1\%$ they reported in the Figure 6 of the main section of the paper refers to the ability of the $\beta$-VAE to provide a feature map with which a classifier can predict whether $x$-position, $y$-position, scale and rotation are shared between pairs of images, while ignoring shape. This is stated in Appendix A.4 of their paper.} We additionally calculated the disentangling metric on the more challenging task of distinguishing between \emph{all 5} of the latent variables.

The results of our experimentation are displayed in Figure~\ref{fig:disentanglement-vs-reconstruction}.
We were able to replicate their results showing that the $\beta$-VAE is capable of achieving essentially $100\%$ on the 4-variable disentanglement task (Figure \ref{subfig:disentanglement-vs-reconstruction-4}), and that good disentanglement of $\beta$-VAE comes at the expense of poorer reconstruction. 
On the 4-variable disentanglement task, we found that WAEs were able to attain similar levels of disentanglement while retaining significantly better reconstruction errors. Note that in this case it is not really possible to get better disentanglement than the $\beta$-VAE, as it already achieves a score approaching $100\%$.
On the 5-variable task (Figure \ref{subfig:disentanglement-vs-reconstruction-5}), WAEs significantly outperformed $\beta$-VAEs simultaneously in terms of disentanglement and reconstruction.

\begin{figure}[t!]
	\centering
	\begin{minipage}{.02\textwidth}
		\vspace{0.6em}	(a) \\ \vspace{-0.1em} (b) \\\vspace{-0.1em} (c) \\ \vspace{-0.3em} (d)
	\end{minipage}
	\begin{minipage}{.45\textwidth}
		\centering
		\includegraphics[width=\columnwidth]{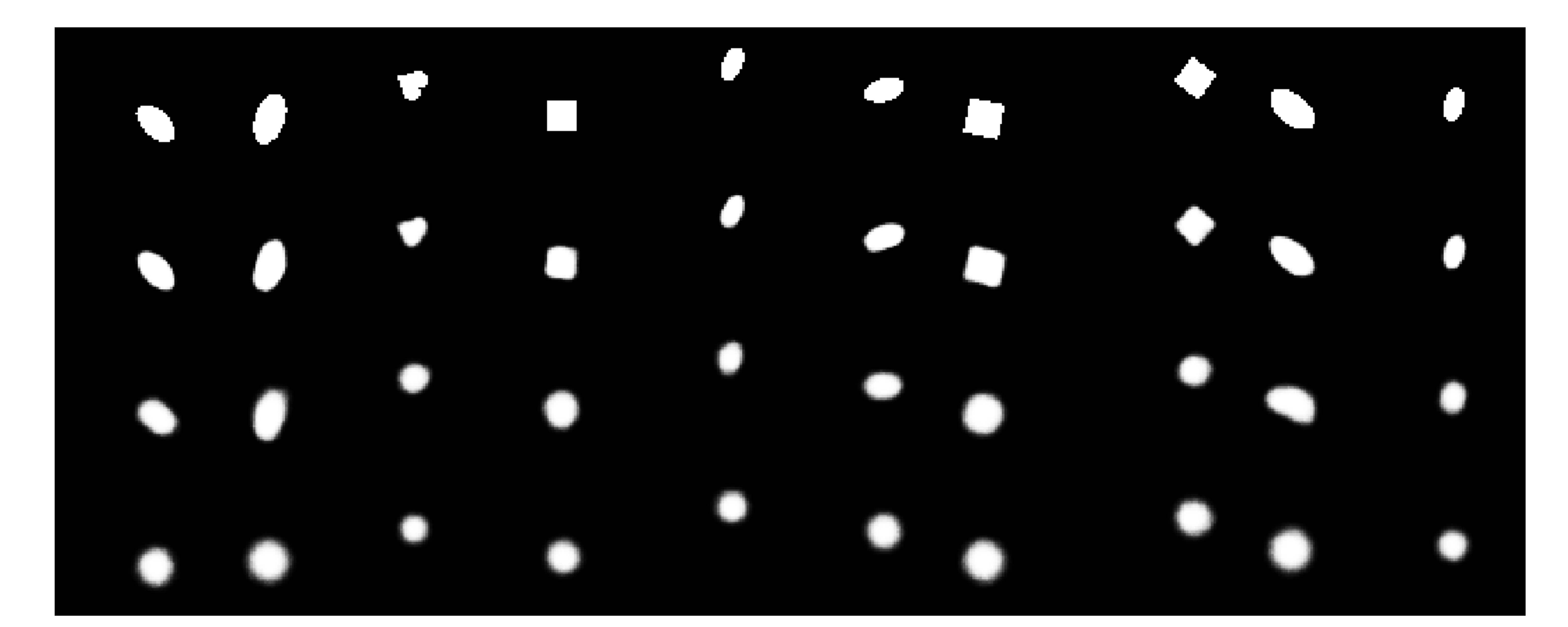} 
	\end{minipage}%
	\caption{\label{fig:dsprites-reconstructions}\textbf{Row (a)}: Samples from the \emph{dSprites} dataset; the remaining rows show reconstructions of these images by: \textbf{Row (b)}: the Gaussian WAE with the best reconstruction error amongst those scoring $>98\%$ on the 4-variable disentanglement metric; \textbf{Row (c)}: the Gaussian WAE with the best score on the 5-variable disentanglement metric; \textbf{Row (d)}: the  $\beta$-VAE with the best reconstruction error amongst those scoring $>98\%$ on the 4-variable disentanglement metric. This visually confirms what is shown in Figure \ref{fig:disentanglement-vs-reconstruction}, namely that WAEs can disentangle better than $\beta$-VAEs while preserving better reconstructions.}
\end{figure}

Amongst all of the $\beta$-VAEs we trained attaining a $4$-variable disentanglement score of $>98\%$, the lowest training reconstruction error was $114.8\pm5.2$. 
The corresponding error for the WAEs was $40.8\pm 1.5$. 
The WAE with the best $5$-variable disentanglement scored an average of $\mathbf{98.8\%}$ across the 3 independent disentanglement calculations for this experiment with a test reconstruction of $94.0\pm5.8$. The $\beta$-VAE performing best on the $5$-variable disentanglement task scored an average of $\mathbf{85.4\%}$ disentanglement with a test reconstruction of $156.1\pm5.7$.
In summary, WAEs are able to outperform $\beta$-VAEs simultaneously in terms of disentanglement metric and reconstruction error.
Sample reconstructions from each of the aforementioned experiments are displayed in Figure \ref{fig:dsprites-reconstructions}.

\section{Conclusion and future directions}

We investigated the problems that can arise when there is a mismatch between the dimension $\dZ$ of the latent space of a WAE and the intrinsic dimension $\dI$ of the dataset on which it is trained. 
We propose to use random encoders rather than deterministic encoders to mitigate these problems.
In practice, we found that additional regularisation on the variances of the encoding distributions was required. 
With this regularisation, random-encoder WAEs are able to adapt to the case that $\dI \ll \dZ$.
We applied regularised random-encoder WAEs to a benchmark disentangled representation learning task on which good performance was observed.

One direction for future research is to investigate whether it is possible for random-encoder WAEs to automatically adapt to $\dZ$ without any hyper-parameter tuning. 
Approaches to this include trying to derive theoretically justified regularisation to prevent variance collapse and to consider new divergence measures that take into account the encoding distribution variances.
The results of our experiments on the disentanglement benchmark combined with the flexibility of the WAE framework indicate that WAEs have the potential to learn useful semantically meaningful representations of data.

\bibliography{wae18icml}
\bibliographystyle{icml2018}

\end{document}